\documentclass{article}

\usepackage{arxiv}

\usepackage{url}            
\usepackage{booktabs}       
\usepackage{amsfonts}       
\usepackage{nicefrac}       
\usepackage{microtype}      
 \usepackage{hyperref}
 \hypersetup{colorlinks=true,linkcolor=citeblue,citecolor=citeblue,urlcolor=citeblue}
\usepackage{cite}
\usepackage{amsmath,amssymb,amsfonts}
\usepackage{cleveref}
\usepackage{subcaption}
\usepackage{makecell}

\usepackage[T1]{fontenc}
\usepackage{lmodern}

\def\*#1*#2{o\null{#2}{#1}}

\def\sh#1{\setbox0=\hbox{#1}%
     \kern-.02em\copy0\kern-\wd0
     \kern.04em\copy0\kern-\wd0
     \kern-.02em\raise.0433em\box0 }

\usepackage{lipsum}
\usepackage{xcolor}
\usepackage{graphicx}
\usepackage{geometry}
\usepackage{array}
\usepackage{multirow}
\usepackage{xcolor}
\definecolor{citeblue}{RGB}{0,0,255}

\graphicspath{ {./images/} }

\title{PotatoGANs: Utilizing Generative Adversarial Networks, Instance Segmentation, and Explainable AI for Enhanced Potato Disease Identification and Classification}

\author{
\\
\newline
\\
Fatema Tuj Johora Faria \textsuperscript{1\textbf{†}*},
Mukaffi Bin Moin \textsuperscript{1\textbf{†}},
Mohammad Shafiul Alam \textsuperscript{1\textbf{†}},
Ahmed Al Wase\textsuperscript{1},
\\
Md. Rabius Sani\textsuperscript{1},
Khan Md Hasib\textsuperscript{2}
\\
\bigskip
\\
\\\textsuperscript{\textbf{1}}Ahsanullah University of Science and Technology, Dhaka, Bangladesh.
\\
\textsuperscript{\textbf{2}}The University of Western Australia, Crawley, WA, Australia.
\\
\\
\bigskip
*Corresponding author(s). E-mail(s): \texttt{\textcolor{blue}{fatema.faria142@gmail.com}}\\
Contributing authors: \texttt{\textcolor{blue}{mukaffi28@gmail.com}}; \texttt{\textcolor{blue}{ahmed.alwasi34@gmail.com}}; \\\texttt{\textcolor{blue}{rshridoy010113@gmail.com}}; \texttt{\textcolor{blue}{shafiul.cse@aust.edu}}; \texttt{\textcolor{blue}{khanmdhasib.aust@gmail.com}}; 
\\ 
\textsuperscript{†}These authors contributed equally to this work.
}

\begin{document}

\maketitle
\abstract{
Numerous applications have resulted from the automation of agricultural disease segmentation using deep learning techniques. However, when applied to new conditions, these applications frequently face the difficulty of overfitting, resulting in lower segmentation performance. In the context of potato farming, where diseases have a large influence on yields, it is critical for the agricultural economy to quickly and properly identify these diseases. Traditional data augmentation approaches, such as rotation, flip, and translation, have limitations and frequently fail to provide strong generalization results. To address these issues, our research employs a novel approach termed as PotatoGANs. In this novel data augmentation approach, two types of Generative Adversarial Networks (GANs) are utilized to generate synthetic potato disease images from healthy potato images. This approach not only expands the dataset but also adds variety, which helps to enhance model generalization. Using the Inception score as a measure, our experiments show the better quality and realisticness of the images created by PotatoGANs, emphasizing their capacity to resemble real disease images closely. The CycleGAN model outperforms the Pix2Pix GAN model in terms of image quality, as evidenced by its higher IS scores CycleGAN achieves higher Inception scores (IS) of 1.2001 and 1.0900 for black scurf and common scab, respectively. This synthetic data can significantly improve the training of large neural networks. It also reduces data collection costs while enhancing data diversity and generalization capabilities. Our work improves interpretability by combining three gradient-based Explainable AI algorithms (GradCAM, GradCAM++, and ScoreCAM) with three distinct CNN architectures (DenseNet169, Resnet152 V2, InceptionResNet V2) for potato disease classification. This comprehensive technique improves interpretability with insightful visualizations and provides detailed insights into the network's decision-making. The goal of combining several CNN designs and explanation techniques is to maximize interpretability while offering an in-depth understanding of the model's behavior. We further employ this extended dataset in conjunction with Detectron2 to segment two classes of potato disease images, with the primary goal of enhancing the overall performance of our model. Furthermore, in the ResNeXt-101 backbone, Detectron2 has a maximum dice score of 0.8112. This combination of PotatoGANs and Detectron2 has the potential to be a powerful approach for tackling the limitations of traditional data augmentation approaches while also boosting the accuracy and robustness of our disease segmentation model.}



\section{Introduction}\label{introlab}
Agriculture is the most important factor in ensuring global food security. It serves as the foundation of sustenance, producing the supplies required to feed the world's ever-increasing population. Agriculture's significance goes beyond mere production; it is an essential component of our capacity to fulfill the nutritional demands of varied cultures across the world \cite{Arshaghi}. Among the huge number of crops grown globally, potatoes stand out as agricultural symbols, playing a critical part in the global agricultural landscape. Beyond their distinctive appearance, potatoes play an important role in global population sustainability. This adaptable crop not only strengthens the nutritional base of numerous regions, but also assures a steady supply of critical components. Potato crops' overall relevance in international agriculture is highlighted by their versatility and nutritional worth, making them an essential component in the goal of global food security \cite{ourPaper}. Furthermore, the cultivation and use of potatoes demonstrate the complicated interplay between agriculture, nutrition, and the vast weave of food systems that define our interdependent world.

Potato diseases, such as notorious Early Blight and Late Blight \cite{Khalifa} \cite{Hou} \cite{tambe} are intense antagonists, posing a serious danger to the global potato agricultural industry. Their stealthy character allows for fast spread, resulting in considerable productivity losses over the whole range of potato-dependent regions. The consequences are most severe in regions where potatoes are a key staple, as these diseases not only damage farming efforts but also disturb the delicate fabric of the larger food supply chain, generating downstream effects across varied agricultural systems. The consequences for the economy are far-reaching and multifaceted, affecting farmers, industry participants, and consumers across several fronts. Farmers suffer the financial burden, dealing with lower yields, rising disease management expenses, and the impending threat of crop failures, making communities largely reliant on potatoes particularly vulnerable to food insecurity \cite{Iqbal}. This financial burden is highlighted in Bangladesh, where potato producers face an annual economic setback of Tk 2,500 crore due to the dual issues of surplus production being unsold and postharvest losses \cite{Ghosh} \cite{Ashraful}. The economic consequences go beyond individual farms and affect the entire business, as rising disease management expenses add to an intricate chain of problems. Industry competitors, including manufacturers and wholesalers, are negotiating an uncertain environment that complicates the already delicate supply-demand balance. Consumers pay additional costs as well, such as potential price increases and limited supply of this crucial product \cite{Singh}. This complicated chain of difficulties highlights the vital need for new and effective methods to manage surplus crops and reduce postharvest waste. Addressing these difficulties front on is critical not only for protecting agricultural livelihoods, but also for building a more sustainable and economically robust potato production business on a worldwide scale, particularly in countries affected hardest by these dangerous potato diseases. The search for answers becomes critical to guaranteeing food security, economic stability, and the overall well-being of communities that are closely related to the success of potato agriculture \cite{Bangari}.

With advances in agricultural technology and the application of artificial intelligence in crop disease diagnosis, it is critical to conduct relevant research to ensure long-term agricultural development. Various diseases caused by viruses and bacteria have a major effect on crop quality and quantity, and manual interpretation of these crop diseases is time-consuming and complicated. Because it takes a high degree of expertise, effective and automated disease identification during the budding phase can help improve crop productivity. Manual methods are time-consuming and yield erroneous results \cite{reddy}. To reduce crop losses, early disease identification and prevention are essential. Automated detection of plant and fruit diseases is an important research issue in the realm of agricultural product monitoring and surveillance \cite{Arshad}. Symptoms of the disease develop on plant leaves as quickly as possible, and millions of liters of toxic chemicals are used each year to manage plant pests, diseases, and weeds in fields. To lessen environmental impact, it's important to use these resources more effectively. In today's world, image-processing applications are growing by the day \cite{Goyal} \cite{Monzurul}. 

In recent research papers, the use of computer vision in agriculture for disease diagnosis has gained popularity. Studies have used computer vision techniques to detect diseases such as late blight in potatoes \cite{Hou}, citrus canker in orchards \cite{ZHANG}, and wheat rust \cite{Shafi}, allowing for early measures to be taken. Automated detection approaches have been applied for tomato bacterial spots \cite{abdulridha}, grapevine leafroll disease \cite{Sawyer}, and maize streak virus \cite{Masood}, adding to precision agriculture and minimizing crop losses. Notably, the impact of disease monitoring in orchards has been improved by the vital role of computer vision in recognizing apple scab signs \cite{Kodors}, according to current study findings. 

Several research papers demonstrate the use of deep learning in potato disease classification. In this research paper \cite{Hou}, the authors used a Kaggle dataset of 1,722 potato leaf photos to address low training data by employing data augmentation methods. These included Reflection X, Reflection Y, Reflection XY, and Zoom, which increased the collection to 9,822 photos. The suggested 14-layer design obtained an amazing 98\% testing accuracy in distinguishing between healthy, early blight, and late blight leaves. Another investigation \cite{Oppenheim1} found that VGG was successful in categorizing potato illnesses with just 400 polluted photos, using mirroring and cropping to construct 2,465 patches. The dataset included a variety of diseases, including Black Dot, Black Scurf, Common Scab, Silver Scurf, and uninfected patches. A related study \cite{Ashraful} described a Customized Convolutional Neural Network (CNN) for diagnosing common potato illnesses, such as Early Blight, Late Blight, and healthy leaves, using a dataset of 10,000 images from Kaggle and other sources. Traditional data augmentation techniques were used, including translations, rotations, scale modifications, shearing, and flipping. The research investigation \cite{InceptionResNetV2}, which focused on potato leaf diseases, employed data collection, pre-processing, augmentation, and image classification to enhance illness detection. The work seeks to increase CNN performance by changing its architecture, resulting in improved accuracy with less trainable parameters. The customized CNN was trained on a dataset of 2,652 pictures uniformly distributed over three classes, specifically targeting potato leaf blight, solving difficulties like accuracy and computation time while emphasizing efficiency. 

Existing research papers have mostly concentrated on potato leaf diseases, emphasizing leaf classification while ignoring the critical challenge of directly classifying whole crops. Unfortunately, current studies followed the usual pattern of ignoring the vital importance of crop classification by focusing just on leaves. This absence is noteworthy, as crop categorization is critical for accomplishing complete disease management in potato cultivation. Furthermore, the studies fail to look into disease localization, which is crucial in understanding where and how potato illnesses occur on the crop as a whole. The lack of disease localization is an impediment limiting insights into disease spreading and preventing the development of detailed and effective disease control strategies for potato cultivation. Bridging this gap is critical for designing comprehensive solutions that go beyond leaf-level research and consider the overall health of the plant. Our research uses direct crop categorization to give a more thorough understanding of disease dynamics in potato fields, allowing for focused solutions and specific disease management approaches.

In this research, we propose a change from typical data augmentation strategies in favor of the use of Generative Adversarial Networks (GANs) as a fundamental component. Our major goal is to use GANs, especially models like CycleGAN and Pix2Pix, to create accurate representations of infected potatoes by converting healthy potato photos into damaged equivalents. Our work focuses on image generation, with the primary objective of effortlessly transitioning healthy data into disease-affected representations. This strategy, which employs cutting-edge GANs, presents a novel methodology for augmenting the dataset and capturing the complex nature of potato diseases in a more realistic and expanded way. 
Furthermore, our secondary goal is to apply Explainable AI (XAI) techniques to improve interpretability and transparency in the potato disease classification procedure. Using XAI approaches, we want to shed light on the classification model's decision-making processes and reveal how it recognizes and differentiates between different potato illnesses. This flexibility helps to build confidence and dependability in the disease categorization findings in addition to improving comprehension of the model's behavior. Moreover, our third goal is sophisticated image segmentation using methods like Detectron 2. This stage is critical for properly finding and distinguishing disease-affected areas within the generated images, providing a better level of detail for disease research.

This study provides an important combination of cutting-edge computer vision and machine learning technology, designed only for agricultural disease detection. We want to increase disease identification and localization accuracy by expanding beyond traditional augmentation approaches. Finally, our study helps to create more successful disease management techniques in agriculture by establishing a strong framework for improved accuracy in diagnosing and understanding potato diseases. \\ 
\\  
 The main points of our contributions to this research study are summarized below:
\begin{itemize}
    \item We propose an innovative GAN approach, the first effort at creating realistic disease images developed primarily for potatoes, expanding the scope of agricultural research.
    \item Our research explores the creation of a unique dataset that includes both generated images and manually annotated samples for exact disease segmentation. The dataset's reliability is further strengthened by verification with the respected Bangladesh Agricultural Research Institute (BARI).
    \item We utilize advanced image generation evaluation metrics, such as Fréchet Inception Distance and Inception Score, to systematically examine the authenticity and quality of the generated disease images, setting a high standard for evaluation methods. 

    \item We enhance interpretability through a fusion of three gradient-based Explainable AI methodologies (GradCAM, GradCAM++, and ScoreCAM) with three diverse CNN architectures (DenseNet169, ResNet152 V2, InceptionResNet V2). This approach delves into the intricacies of the network's decision-making processes, providing a deeper understanding through insightful visualizations.

    \item Our research significantly improves potato disease segmentation by utilizing Detetron 2, a cutting-edge image segmentation tool. We use modern evaluation matrices such as Intersection over Union and Dice Coefficient to ensure an accurate evaluation of the accuracy and efficacy of our segmentation processes.
\end{itemize}

The remaining content of the paper is organized as follows: Section \ref{Relatedlab} provides an in-depth review of relevant literature, establishing a foundation for our research. Section \ref{Datasetlab} explores every detail of our dataset creation process. Section \ref{Backgroundlab} explores background studies regarding Generative Adversarial Networks, Convolutional Neural Networks, Explainable AI techniques,  Instance Segmentation Model, and Evaluation Metrics. Section \ref{Methodologylab} explores the proposed methodology, outlining the analytical steps and reasoning behind our research objectives. Section \ref{resultlab} provides a detailed discussion of the experimental results. Section \ref{futurelab} discusses future research prospects and a roadmap for enhancing precision agriculture. Section \ref{conclusionlab} summarizes the key findings of our study and their significance for potato disease prevention and management in potato agriculture. 


\section{Related Works} \label{Relatedlab}

In this comprehensive literature review, we systematically investigate three critical factors in crop disease detection. The first component \cref{sec: Classification} entails a detailed examination of existing research and approaches for crop disease detection using image classification. By analyzing a variety of techniques, we want to provide a comprehensive summary of developments in this field. The second category \cref{sec: GANS} delves into the domain of Generative Adversarial Networks (GANs) in Crop Disease Detection, where we look at GANs' remarkable ability to generate synthetic data and explore their applications in improving crop disease detection accuracy and efficiency. The third and final component \cref{sec: Segmentation} focuses on Image Segmentation Techniques for Crop Disease Detection, highlighting how important these methods are in accurately recognizing and localizing unhealthy spots in crop images. Our comprehensive literature review aims to give a detailed perspective of the cutting-edge approaches, obstacles faced, and potential future directions in the changing environment of crop disease detection. Table \ref{relatedtable} beautifully summarizes previous research, providing a brief and organized overview of major results and contributions in the field.

\subsection{\textbf{Image Classification Techniques for Crop Disease Detection}}\label{sec: Classification}

Arshaghi et al. \cite{Arshaghi} focused on the identification and classification of surface defects in potatoes using image processing approaches and Convolutional Neural Networks (CNN), such as AlexNet, GoogLeNet, VGG, R-CNN, and Transfer Learning. The research was performed using a dataset of 5000 potato photos. The CNN model, fine-tuned by altering the number of layers, performed exceptionally well, outperforming other models with substantial accuracy. In several classes, the CNN model obtained outstanding accuracy rates of 100\% and 99\%, demonstrating its usefulness in precisely identifying and classifying potato surface flaws. In a separate study, Oppenheim et al.\cite{Oppenheim1} conducted another study on potato illness classification using Convolutional Neural Networks. The study relied on a large dataset of 2,465 diseased potato patches. The study discovered a striking fact while implementing various Train-Test Set Divisions: increasing the data given to the training phase resulted in a significant improvement in classification performance. This improvement was accompanied by a significant reduction in error rates, emphasizing the importance of dataset quantity in developing CNN-based potato disease classification models. Mahum et al. \cite{Mahum} focused on potato leaf disease classification, using both ``The Plant Village'' dataset and manually obtained data. The study used a modified pre-trained DenseNet-201 model with an extra transition layer to improve illness classification efficiency. This technique produced impressive results, with a remarkable accuracy of 97.2\%, demonstrating the modified DenseNet-201's efficiency in reliably identifying and classifying potato leaf diseases. In another research, Faria et al. \cite{ourPaper} proposed a hybrid strategy for potato disease classification that combines image processing with MobileNet V2, LSTM, GRU, and BiLSTM. The MobileNet V2-GRU setup, which was optimized utilizing Stochastic Gradient Descent, performed exceptionally well, reaching 99\% accuracy on the test dataset. This study emphasized combining several designs to improve accuracy in agricultural disease classification. The success of MobileNet V2-GRU demonstrated the positive effects of merging lightweight convolutional and recurrent neural networks. This research made a significant addition to optimizing potato disease classification approaches, stressing the practical usefulness of the hybrid model in the real world.

\subsection{\textbf{Generative Adversarial Networks (GANs) in Crop Disease Detection}} \label{sec: GANS}
Yilma et al. \cite{Getinet} developed a novel method for detecting and classifying tomato diseases using the Attention Augmented Residual (AAR) network. This network combines a residual block that has been pre-activated for coarse-level feature learning with an attention block that captures important features and global associations. To improve the training dataset, they also used a Conditional Variational Generative Adversarial Network (CVGAN) for image reconstruction and augmentation approaches. The AAR network demonstrated outstanding accuracy in trials, with scores ranging from 97.04\% to 99.03\%, regularly exceeding typical CNN models. Furthermore, the Complex-Wavelet Structural resemblance Index (CW-SSIM) was used in the study to analyze the quality of produced samples by evaluating their resemblance to the ground truth for each class. The CVGAN-generated samples were 87.53\% comparable to the original training samples on average. In another research, Zhao et al. \cite{Zhao} introduced DoubleGAN, a two-stage approach to generate high-resolution plant leaf images. In the first stage, healthy and diseased leaf images are utilized to train a Wasserstein Generative Adversarial Network (WGAN) to generate a pretrained model. Following that, the pretrained model is used to create 64x64 pixel images of unhealthy leaves. In the second stage, a Super-Resolution Generative Adversarial Network (SRGAN) is used to upscale the 64x64 pictures to 256x256 pixels, giving extensive details. The results show that images created by DoubleGAN are clearer than those produced by Deep Convolutional Generative Adversarial Network (DCGAN). The accuracy of plant species and disease recognition is 99.80\% and 99.53\%, respectively. Furthermore, the study applies three traditional networks for classification (VGG16, ResNet50, and DenseNet121), proving the similarity of DoubleGAN-generated sick plant leaf pictures to legitimate photos. The achieved accuracy rates for plant species identity and disease recognition are 99.80\% and 99.53\%, respectively. Cap et al. \cite{Cap} present LeafGAN, an innovative image-to-image translation system with a distinct attention mechanism meant to diversify diseased images for improved plant disease identification, in their study. LeafGAN outperforms vanilla CycleGAN in data augmentation, improving diagnostic accuracy by 7.4\%, a significant improvement when compared to CycleGAN's minimal 0.7\% improvement. LeafGAN focuses on producing realistic disease images from healthy equivalents while keeping various backgrounds, emphasizing its significance as an excellent data augmentation tool for practical plant disease detection. The study also emphasizes the importance of the innovative Label-Free Leaf Segmentation (LFLSeg) module, which achieved an outstanding 99.8\% accuracy in categorizing various leaf objects. Although LFL-Seg performs somewhat worse than the AOP network with pixel-level labeled training images, it is dependable without masking training data. The large changes in image integration with the segmentation mask, resulting in improved classification performance, highlight the usefulness of LeafGAN, directed by LFLSeg to alter just the leaf area. In a separate study, Ramadan et al. \cite{Ramadan} proposed an innovative method for improving maize leaf disease classification. They utilized synthetic pictures created by CycleGAN to augment the dataset used to train deep learning models, such as ResNet50V2, DenseNet169, VGG16, VGG19, Xception, MobileNetV2, ViT-B/16, and ViT-B/32. Notably, DenseNet169 outperformed the other models, reaching 98.48\% accuracy when trained on the CycleGAN-enhanced dataset. This study demonstrated the efficiency of using synthetic data with advanced deep-learning architectures to greatly enhance agricultural disease categorization.

\subsection{\textbf{Image Segmentation Techniques for Crop Disease Detection}}\label{sec: Segmentation}

Afzaal et al. \cite{Afzaal} bring an important contribution to the area of plant disease identification by offering a unique dataset of 2500 images including seven forms of strawberry diseases. This dataset is an important resource for the development of deep learning-based autonomous detection systems, notably for segmenting strawberry diseases in the presence of complicated background conditions. Various augmentation strategies, such as changes to geometry, color, and arithmetic, were used to improve the dataset, led by an optimal augmentation graph. The research focuses on using the Mask R-CNN architecture to effectively separate instances of the detected strawberry diseases. The employment of a ResNet backbone combined with a systematic approach to data augmentation offers encouraging results, with a mean average precision (mAP) of 82.43\%. The complete evaluation includes experiments with ResNet50 and ResNet101 backbones, both initialized with pre-trained MS-COCO Res-Net101 weights. Fu et al. \cite{Fu} presented a unique strategy for potato leaf disease spot segmentation in their research by adopting an improved UNet network model. To address issues such as gradient vanishing and degradation, the methodology includes ResNet50 as the backbone network. It acts as the decoder, successfully combining potato-specific features by utilizing the distinctive characteristics of the UNet network. Furthermore, the study employs the SE (squeeze and excitation) attention mechanism on ResNet50 to improve the network's capacity to acquire disease spot features by highlighting essential input while suppressing irrelevant ones.  They evaluate the model's effectiveness using a wide range of measures, such as the Dice coefficient, intersection over union (IoU), and accuracy (Acc). The careful training method includes a batch size of 1 and 300 iterations of optimization using the stochastic gradient descent (SGD) optimization. The results show that the RS-UNet model outperforms the classic UNet technique, with an excellent Dice coefficient of 88.86\%. Li et al. \cite{Li} suggested a three-stage integrated framework for the segmentation and detection of potato foliage diseases in complex backgrounds in their study, which combined instance segmentation, classification, and semantic segmentation models. The first stage uses Mask R-CNN to accurately separate potato leaves in complicated backgrounds, with an average precision (AP) of 81.87\% and a precision of 97.13\%. In the second stage, the classification models VGG16, ResNet50, and InceptionV3 classify potato leaves with accuracies of 95.33\%, 94.67\%, and 93.33\%, respectively.  The third stage identifies and detects disease areas using UNet, PSPNet, and DeepLabV3+ semantic segmentation models, with mean intersection over union (MIoU) values of 89.91\% and mean pixel accuracy (MPA) values of 94.24\%. The three-stage methodology lowers the impact of wild conditions on potato leaf segmentation, enhances disease spot segmentation accuracy, and offers technological assistance for potato leaf disease diagnosis and prevention. In another research, Rashid et al. \cite{Rashid} developed a new multi-level deep learning model for potato leaf disease analysis, which included YOLOv5 for leaf extraction and a unique Potato Leaf Disease Detection Convolutional Neural Network (PDDCNN) for disease classification. Notably, the PDDCNN method outperformed other techniques on both the PlantVillage and PLD datasets, exhibiting higher performance even in cross-dataset comparisons. In direct contrast to existing methodologies, PDDCNN obtained an impressive 99.75\% accuracy, demonstrating its efficacy as an improved solution for accurate and robust potato leaf disease detection.

\begin{table}
\caption{Overview of Key Findings in Image Classification Techniques, Generative Adversarial Networks (GANs), and Image Segmentation Techniques for Crop Disease Detection.}\label{relatedtable}
  \begin{tabular}{p{2.0cm} p{1.8cm} p{1.8cm} p{2.3cm} p{8.0cm}} 
    \toprule
 \textbf{Type} & 
 \textbf{Reference} &
\textbf{Publication \newline Year}&
\textbf{Models \newline Employed}&
\textbf{\textbf{Contribution}} \\
    \midrule
  \textbf{Classification}  & \cite{Arshaghi}  & 2022 & CNN, Alexnet, Googlenet, VGG, R-CNN & The CNN model was optimized by configuring the number of layers and it achieved remarkable accuracy outperforming other models, hitting 100\% and 99\% accuracy in some of the classes respectively. \\ \\
   & \cite{Oppenheim1}  & 2017 & CNN & The CNN attained a robust 95.85\% accuracy in classifying diverse potato diseases from 2,465 images, showcasing adaptability to varied acquisition conditions.  \\ \\
     & \cite{Mahum}  & 2022 & DenseNet-201 &  Employed a modified pre-trained DenseNet-201 model with an additional transition layer for efficient disease classification. This approach achieved a 97.2\% accuracy. \\ \\
     & \cite{ourPaper}  & 2023 & MobileNet V2, LSTM, GRU, BiLSTM & A hybrid approach using image processing combined MobileNet V2 with LSTM, GRU, and Bidirectional LSTM for potato disease classification. MobileNet V2-GRU with Stochastic Gradient Descent achieved 99\% accuracy.  \\
    \midrule
      \textbf{Generation} & \cite{Getinet}  & 2021 & AAR network, CVGAN & A CVGAN reconstruction network and augmentation techniques were applied to handle the small amount of training data.\\ \\
     & \cite{Zhao}  & 2021 & DoubleGAN, WGAN, SRGAN, DCGAN & DoubleGAN addressed imbalances in plant leaf datasets by generating high-resolution images of unhealthy leaves with fewer samples. It outperformed DCGAN, producing clearer images that surpassed the original dataset results.  \\ \\
     & \cite{Cap}  & 2020 & LeafGAN, Vanilla CycleGAN & LeafGAN enhanced the robustness of a cucumber plant disease diagnostic system by serving as an effective data augmentation method, producing visually superior images compared to vanilla CycleGAN.  \\ \\
     & \cite{Ramadan}  & 2023 & CycleGAN & CycleGAN was employed to generate synthetic images, enhancing maize leaf disease classification.\\
    \midrule
    \textbf{Segmentation} & \cite{Fu}  & 2023 & RS-UNet, ResNet50 & Introduced RS-UNet for accurate segmentation and localization of potato leaf disease spots in photos, using a combination of cross-entropy and dice loss.\\ \\
     & \cite{Li}  & 2022 & Mask R-CNN, UNet, PSPNet, and DeepLabV3+ & Combined Mask R-CNN with VGG16, ResNet50, and InceptionV3 for classification and UNet, PSPNet, and DeepLabV3+ to enhance semantic segmentation capabilities.  \\ \\
     & \cite{Afzaal}  & 2021 & Mask R-CNN  & A Mask R-CNN model with a ResNet backbone and systematic data augmentation efficiently executed instance segmentation for diseases.  \\ \\
     & \cite{Rashid}  & 2021 & YOLOv5 & Used YOLOv5  for leaf extraction and a unique potato leaf disease detection \\
    \bottomrule
  \end{tabular}
\end{table}

\section{Dataset Creation} \label{Datasetlab}

This section explores the details of Dataset Creation, explaining the laborious method of creating our dataset from the bottom up. It provides a thorough explanation of the methods used to create an extensive and useful dataset suited to the specific needs of our research.
    \subsection{ \textbf{Data Collection}}
    We begin by collecting a variety of images of healthy potatoes. These images show the potatoes from different angles, in varied lighting conditions, and at different stages of growth. We create a separate collection for our dataset of diseased potatoes. This collection comprises two diseases with varying degrees of severity: Black Scurf and Common Scab. The images are gathered from several agriculture websites. We have 93 images of Black Scurf and an additional 126  images of Common Scab in our collection, which has helped to improve our dataset.

\begin{table}
    \caption{Overview of Dataset Specifications}
    \label{table2}
    \centering
    \begin{tabular}{p{3.0cm} p{3.8cm} p{3.2cm}p{1.8cm}}
        \toprule
        \textbf{Disease Name} & \textbf{Disease Symptoms} &  \textbf{Causes of Disease} & \textbf{Number of Images}\\
        \midrule
        Common scab & Raised, corky lesions or scabs manifest on the surface of potato tubers & Bacteria & 126 \\ 
        Black scurf &  Dark, irregular lesions develop on the surface of potato tubers & Fungus & 93 \\
    \bottomrule
    \end{tabular}
\end{table}

    \subsection{ \textbf{Disease \& Healthy Potato Image Labeling}}
        During this stage, we manually label all of the images in our dataset to differentiate between ``healthy'' potatoes and those impacted by a disease. Our dataset, which included two types of disease-infected potato images, was validated by the Bangladesh Agricultural Research Institute (BARI). Our categorization procedure for healthy potato images was done manually by human experts, who considered outward looks as well as specific qualities like color, shape, and texture. This hands-on labeling strategy assures that our dataset is precise and reliable for future analysis and model training.

    \subsection{ \textbf{Image Preprocessing}} 
We begin by standardizing and improving the visual quality of our dataset as the first step in our thorough image processing for detailed analysis of potato disease photos. To provide a foundation for uniformity in further studies, the first essential step is to resize each image to a constant resolution of (\(224 \times 224\)) pixels. The noise in the potato disease images is then addressed and reduced by carefully using an advanced bilateral filter. This sophisticated filtering method not only makes the photos clearer, but also adds a smoothness that makes the impacted potato samples seem much better overall. A key step that lays the foundation for more accurate, precise, and trustworthy studies of the numerous illnesses affecting potatoes is the reduction of noise using the bilateral filter. The next step in this transformation process is to use contrast stretching, which is a method that goes beyond simple noise reduction and resizing. To bring out the subtler features and nuances present in the symptoms, this stage explores the complex world of boosting contrast and brightness levels inside every image of potato disease. By using this approach, the entire visual experience is improved and small variations in how the disease presents itself are suitably brought to light. Apart from these fundamental techniques, we also carefully use CLAHE (Contrast Limited Adaptive Histogram Equalization), Random Brightness, and HSL (Hue, Saturation, and Lightness) adjustments. More specifically, CLAHE serves as an effective instrument to dynamically boost the contrast of our potato disease images, ensuring an even improvement in all impacted areas. These additional techniques work as precise strokes of paint, refining the potato disease image gathering with a keen perception of visual richness and coherence. After all of these digital image processing stages are completed, a dataset is produced that goes beyond simple standardization. It turns into a well-chosen collection of images showing potato diseases, with each pixel carefully chosen and adjusted to make sense as a whole. In addition to being ready for additional examination, this visually appealing dataset is proof of the great care that was taken at every stage of its conversion, improving our knowledge of potato diseases for more precise study and diagnosis.

    \subsection{ \textbf{Data Augmentation}}
We utilize a number of advanced techniques in our systematic approach to improving the variety and flexibility of our dataset for model training. By exposing our models to a wide range of circumstances, these techniques aim to develop their competency and durability. First, we present the idea of picture flipping, which is similar to looking into a mirror. By flipping photos horizontally, this transformational approach gives our models a new point of view. This technique trains our models to recognize things from diverse orientations, much as we may examine an item from different perspectives. It's a vital phase in helping them comprehend more and become better at handling other points of view. We extend our approach of dataset augmentation to include vertical flipping, which goes beyond simple reflection. This method adds even more diversity to the dataset by inverting an item. We guarantee a more thorough training experience for our models and increase their flexibility by exposing them to images that have been flipped both along the vertical and horizontal axis. As we rotate images in an organized way to produce orientation changes, rotation becomes an additional vital part of our approach. This intentional difference adds to the overall resilience of our models by strengthening them and enabling them to handle objects at varying angles. To achieve real-world flexibility, we include brightness level changes between images. By simulating different lighting conditions, this dynamic adjustment helps our models be ready for the unpredictability of real-world scenarios. We improve their capacity to properly generalize by exposing them to a variety of lighting settings. To provide even more variation to our collection, we make small color changes to potato images. Potato color varies somewhat as a result of this tiny change, adding to our detailed understanding of color variations. This intentional incorporation of color variation makes our models better capable of handling the many shades of potatoes that they would come across in the real world.
    \subsection{ \textbf{Dataset Splitting}}
    We divide our data into three unique sets, which correspond to our three datasets: one for healthy potato images and two for Black Scurf and Common Scab. Each dataset is split into 80\% for training and 20\% for testing. This intended divide ensures that we have enough data for training our models while providing plenty for testing. This method enables us to reliably test our models' performance on previously unseen data while keeping a balanced representation of healthy and disease-labeled images inside each split.

    \subsection{ \textbf{Image Pairing}}
    After applying augmentation and image processing techniques, our dataset initially comprised 930 instances of Common Scab and 1260 instances of Black Scurf. During the image pairing procedure, we create pairs of images that will be used to train our Pix2Pix GAN model. Each pair has a healthy potato image and a disease-labeled potato image. These paired images serve a specific function: the healthy potato image serves as the ``input'', while the diseased potato image serves as the ``target.'' This coupling enables our Pix2Pix GAN models to comprehend and learn the link between a healthy potato and the disease that it may acquire. We use an approach in which each disease-labeled potato image is coupled with 10 unique healthy potato images to enable complete and strong learning. This systematic coupling exposes the Pix2Pix GAN models to a diverse range of healthful potato shapes and forms. As a result, the models obtain an in-depth understanding of the range of healthful potato looks. When creating disease images, this approach is crucial since it guarantees that the model can account for and create disease-like characteristics over a wide range of healthy potato shapes.  As our CycleGAN operates on unpaired images, image pairing is not required for training, making it a useful model for turning one kind of healthy potato image into another type of diseased potato image without the need for one-to-one image matching.

    \subsection{ \textbf{Data Annotation for Segmentation}}
    After utilizing two GAN models to generate Realistic Disease Images, we carefully selected 1000 of them for segmentation purposes. Following a careful Human Verification process, the final tally revealed 548 instances of Black Scurf and 452 instances of Common Scab among the selected images. We use a thorough manual annotation procedure for our disease potato images in preparation for taking advantage of Detectron2's superior segmentation capabilities. Our goal is to properly indicate the disease-affected areas, allowing our model to recognize and discriminate these areas efficiently. The technique of annotation we use depends on the type of disease. We use pixel-level annotation to ensure proper boundaries for this purpose. We use the VGG image annotator tool, a specialist tool built for image annotation, to help with this procedure.

\section{Background Study} \label{Backgroundlab}
\subsection{Generative Adversarial Networks}
\subsubsection{Pix2Pix GAN}
Pix2Pix \cite{pix2pix} is a specialized type of Generative Adversarial Network (GAN) designed for image translation tasks. It takes paired images for training, with each pair comprising an input image and its corresponding target image. The model contains a generator, responsible for transforming input images into realistic target images, and a discriminator, which determines the generated images against real target images. Through adversarial training, the generator seeks to generate high-quality output images that are distinct from real targets, making Pix2Pix suitable for tasks such as image denoising, colorization, and style transfer. The training process minimizes a combined loss consisting of adversarial loss, which compels the generator to generate images that are indistinguishable from real ones, and pixel-wise L1 loss, which ensures that the generated and target images are similar in terms of pixel values. The representation of the Pix2Pix loss functions are in Equation \ref{Eq:1}, \ref{Eq:2} and \ref{Eq:3}. \\ \\
\textbf{Adversarial Loss (GAN Loss):}
\begin{equation}\label{Eq:1}
\begin{aligned}
   \mathcal{L}_{\text{GAN}}(G, D) = &\mathbb{E}_{x,y}[\log(D(x, y))] \\
   &+ \mathbb{E}_{x,z}[\log(1 - D(x, G(x, z)))]
\end{aligned}
\end{equation}

Here, \(x\) is the input image, \(y\) is the corresponding target image, \(G\) is the generator, \(D\) is the discriminator, and \(z\) is a random noise vector.\\ \\
\textbf{Pixel-wise L1 Loss:}
\begin{equation}
\label{Eq:2}
\mathcal{L}_{L1}(G) = \mathbb{E}_{x,y,z}[\|y - G(x, z)\|_1]
\end{equation}

Here, \(\|y - G(x, z)\|_1\) represents the L1 norm (sum of absolute differences) between the target image \(y\) and the generated image \(G(x, z)\).

The total loss used during training is often a weighted sum of the adversarial loss and the pixel-wise L1 loss:

\begin{equation}
\label{Eq:3}
\mathcal{L}_{\text{total}}(G, D) = \mathcal{L}_{\text{GAN}}(G, D) + \lambda \cdot \mathcal{L}_{L1}(G)
\end{equation}

Here, \(\lambda\) is a hyperparameter that controls the trade-off between the adversarial and L1 components. It determines the relative importance of each term in the total loss. The choice of \(\lambda\) depends on the specific task and the characteristics of the dataset being used for training.

\subsubsection{CycleGAN} 
CycleGAN \cite{cyclegan} is a specialized form of Generative Adversarial Network (GAN) designed for unpaired image-to-image translation tasks, allowing flexible conversion between different domains without the need for one-to-one correspondences in the training data. It employs two generator networks, each dedicated to a specific domain conversion, utilizing convolutional neural networks (CNNs) to learn mappings between domains. Adversarial training is incorporated, where generators aim to produce images indistinguishable from real ones in their respective domains, with discriminator networks providing adversarial feedback. A unique feature is the inclusion of cycle consistency loss, ensuring that translations between domains are consistent by reconstructing the original image after a round-trip conversion, enhancing the overall quality and realism of generated images. 

The loss function of CycleGAN involves a combination of adversarial loss and cycle-consistency loss.
\\
\textbf{ \\ Adversarial Loss:}
\begin{itemize}
    \item Two generative adversarial networks (GANs), one for each domain, are used by CycleGAN.
    \item The adversarial loss is used to train these generators to create images that are identical to real images in the target domain.
    \item To define the adversarial loss, a discriminator is used for every domain. Reducing the likelihood that the discriminator can correctly discriminate between actual and generated images is the generator's objective.
\end{itemize}
   
   For the generator \(G\) translating from domain \(X\) to domain \(Y\), the adversarial loss is:

\begin{equation}
\begin{aligned}
   \mathcal{L}_{\text{GAN}}(G, D_Y, X, Y) = &\mathbb{E}_{y \sim p_{\text{data}}(y)}[\log D_Y(y)] \\
   &+ \mathbb{E}_{x \sim p_{\text{data}}(x)}[\log(1 - D_Y(G(x)))]
\end{aligned}
\end{equation}

   Here, \(D_Y\) is the discriminator for domain \(Y\), and \(G(x)\) represents the generated image from \(x\) in domain \(X\).
\\ 
\textbf{\\ Cycle-Consistency Loss:}
\begin{itemize}
   \item The cycle-consistency loss enforces that translating an image from domain \(X\) to domain \(Y\) and then back to domain \(X\) should yield the original image. The same applies to translations from \(Y\) to \(X\) and back to \(Y\).
    \item This loss helps maintain consistency and ensures that the generated images are semantically meaningful.
\end{itemize}
   The cycle-consistency loss is given by:

\begin{equation}
\begin{aligned}
   \mathcal{L}_{\text{cyc}}(G, F, X, Y) = &\mathbb{E}_{x \sim p_{\text{data}}(x)}[\|F(G(x)) - x\|_1] \\
   &+ \mathbb{E}_{y \sim p_{\text{data}}(y)}[\|G(F(y)) - y\|_1]
\end{aligned}
\end{equation}

   Here, \(F\) is the generator for the reverse translation. \\

After translating from \(X\) to \(Y\), the generator \(G\) aims to minimize both the adversarial loss and the cycle-consistency loss:
\begin{equation}
\begin{aligned}
   \mathcal{L}(G, F, D_X, D_Y, X, Y) = &\mathcal{L}_{\text{GAN}}(G, D_Y, X, Y) \\
   &+ \lambda \cdot \mathcal{L}_{\text{cyc}}(G, F, X, Y)
\end{aligned}
\end{equation}

where \(\lambda\) is a weighting parameter that balances the importance of the adversarial loss and the cycle-consistency loss.

Similar objectives are defined for the generator \(F\) translating from \(Y\) to \(X\), and corresponding adversarial and cycle-consistency losses are calculated for the discriminators \(D_X\) and \(D_Y\).

In summary, the CycleGAN objective is a combination of adversarial training to generate realistic images and cycle consistency to ensure consistency between the input and the translated-reconstructed images.

\subsection{Convolutional Neural Networks}
\subsubsection{DenseNet169}
DenseNet169 \cite{DenseNet169} is a variant of DenseNet, a Densely Connected Convolutional Network. It features dense blocks, enabling each layer to receive input from all preceding layers, promoting feature reuse for more compact representations. The architecture includes bottleneck layers (1x1 convolutions) to reduce input channels and computational complexity. Transition blocks, with 1x1 convolutions followed by pooling, are inserted between dense blocks to reduce spatial dimensions. Global average pooling, employed at the network's end, replaces fully connected layers, reducing parameters for improved generalization. 
\subsubsection{Resnet152V2}
ResNet152v2 \cite{Resnet152V2} is a variant of ResNet, a deep neural network architecture with 152 layers. Notable features include the use of residual blocks, containing convolutional layers with shortcut connections that skip one or more layers, facilitating the training of extremely deep networks. The architecture employs a bottleneck design, incorporating 1x1, 3x3, and 1x1 convolutional layers within each residual block to reduce and restore channel dimensions, optimizing computational efficiency. Batch normalization is applied after each convolutional layer, enhancing training stability. Global average pooling replaces fully connected layers at the network's end, reducing parameters and improving generalization. Pre-activation residual blocks, where batch normalization and ReLU activation precede convolutional layers, contribute to enhanced training performance.

\subsubsection{InceptionResNetV2}
InceptionResNetV2 \cite{InceptionResNetV2} is an extension of the Inception architecture, blending it with residual connections inspired by ResNet. It employs Inception blocks with diverse convolutional filters and pooling to capture features at multiple scales. Residual connections, akin to ResNet, mitigate the vanishing gradient problem for effective training of deep networks. Bottleneck blocks reduce and restore channel dimensions, enhancing computational efficiency. The inclusion of auxiliary classifiers provides intermediate supervision, aiding gradient flow and convergence. Batch normalization stabilizes training, and reduction blocks manage spatial dimensions, collectively making InceptionResNetV2 a powerful architecture for diverse computer vision tasks.
\subsection{Gradient-based Explainable AI}
\subsubsection{GradCAM}
Gradient-weighted Class Activation Mapping or Grad-CAM \cite{GradCAM} is a method for interpreting CNN decisions in challenges with image classification. An input image is fed through a CNN in a feedforward pass, which produces final feature maps that capture hierarchical representations. Backpropagation is used to generate gradients of the predicted class score with respect to the feature maps of the final convolutional layer, which show how modifications affect the final result. These gradients are subjected to global average pooling, which yields weighted averages that emphasize the significance of each feature map. Next, a weighted summation of the feature maps is computed, emphasizing the ones that are most relevant to the predicted class. A heatmap is produced by upsampling and using a ReLU activation function, which highlights positive contributions. This heatmap overlays the original input image, highlighting regions that are critical to CNN's decision-making process and providing a visual representation of the model's decision-making process. Grad-CAM provides information on the regions of the image that affect the classification result.

\subsubsection{GradCAM++}
GradCAM++ (Gradient-weighted Class Activation Mapping++) \cite{GradCAM++} extends the GradCAM method to enhance the interpretability of CNNs in image classification. It starts with a feedforward pass, processing an input image through the CNN to obtain final feature maps. Gradients of the predicted class score are then computed with respect to these feature maps, revealing their impact on the final decision. GradCAM++ introduces a positive-gradient ReLU, considering only positive gradients to emphasize meaningful patterns and discard irrelevant information. Further, it squares the positive gradients, amplifying their importance for better localization. Global Average Pooling is applied to produce weights for each feature map, and a weighted sum of feature maps is computed. A ReLU activation focuses on positive contributions, and the resulting activation map is upsampled to generate the GradCAM++ heatmap, highlighting regions crucial for CNN's prediction. This refined approach provides improved visualization and understanding of the model's decision-making process in image classification tasks.

\subsubsection{ScoreCAM}
Score-CAM (Score-weighted Class Activation Mapping) \cite{ScoreCAM} is a technique for interpreting CNN decisions in image classification. It begins with a feedforward pass, processing an input image through the CNN to obtain final feature maps. The class activation map (CAM) is computed via global average pooling of the feature maps, creating a spatial map highlighting regions relevant to the predicted class. The CAM is then weighted by the class score of the predicted class, emphasizing more relevant regions. A ReLU activation focuses on positive contributions, and the resulting activation map is upsampled to the input image size, forming the Score-CAM heatmap. This heatmap effectively highlights the critical regions influencing CNN's decision for the specific class, offering improved interpretability compared to some other visualization methods.

\subsection{Instance Segmentation Model}
\subsubsection{Detectron 2}  
Detectron 2\cite{detectron2}, developed by Facebook AI Research (FAIR), is a powerful deep-learning framework tailored for object detection and instance segmentation tasks. Evolving from its predecessor, Detectron, Detectron 2 provides a modular and flexible platform for training and deploying computer vision models. It employs well-established architectures like Faster R-CNN, Mask R-CNN, and RetinaNet, utilizing pre-trained convolutional neural networks (CNNs) for a robust foundation. Users can input labeled training data in various formats, allowing adaptability to diverse sources. During training, the framework fine-tunes pre-trained models using stochastic gradient descent and enables customization of hyperparameters for optimal performance. After training, Detectron 2 facilitates efficient inference, generating predictions for object detection and instance segmentation, with additional functionalities for result visualization and interpretation. Its versatility and comprehensive features make it a valuable tool for a wide range of computer vision applications.

\subsection{Evaluation Metrics}
\subsubsection{Evaluation Metrics for Image Generation \newline}

\textbf{The Fréchet Inception Distance (FID)} \cite{FID} is used to evaluate the standard of pictures generated from generative models, particularly in the context of Generative Adversarial Networks (GANs). It is intended to evaluate the similarities of the spatial distribution of real and generated images.

    The formula for the Fréchet distance is as follows:

   \begin{equation} \text{FID} = \left\lVert \mu_r - \mu_g \right\rVert^2 + \text{Tr}(\Sigma_r + \Sigma_g - 2(\Sigma_r \Sigma_g)^{1/2}) \end{equation}
    \newline 
Here: 
\begin{itemize}
    \item \(\mu_r\) and \(\mu_g\) are the means of the real and generated image feature representations.

    \item \(\Sigma_r\) and \(\Sigma_g \) are the covariances of the real and generated image feature representations.
    \item Tr is the trace operator.
\end{itemize}

\textbf{The Inception Score (IS)} \cite{IS} is a metric used to evaluate the quality of generated images from generative models, particularly in the context of Generative Adversarial Networks (GANs). It evaluates the diversity and quality of generated pictures by taking into account both the classifiability and diversity of generated samples.
   Calculate the Inception Score using the formula:
   \begin{equation}
     \text{IS} = \exp\left(\mathbb{E}_{x \sim p_{\text{data}}} \left[ D_{\text{KL}}(p(y | x) || p(y)) \right] \right)   
   \end{equation}
  
    Here: 
\begin{itemize}
    \item \((p(y | x)\) is the conditional class distribution given an image.

    \item \(p(y)\) is the marginal class distribution over all images.
    \item \(D_{\text{KL}}\) is the conditional class distribution given an image.
\end{itemize}

\subsubsection{Evaluation Metrics for Image Segmentation\newline } 

 \textbf{Intersection over Union (IoU)} \cite{IOU} is a measure widely employed in the comparison of segmentation models, notably in computer vision tasks involving the identification and delineation of objects or areas within an image. The overlap between the expected segmentation and the ground truth segmentation is measured by IoU.
    The IoU is calculated using the following formula:
   \begin{equation} \text{IoU} = \frac{\text{Area of Intersection}}{\text{Area of Union}} \end{equation}
\textbf{The Dice Coefficient,} \cite{dice} often known as the F1 score, is a popular statistic for evaluating segmentation models. It calculates the degree of similarity between the anticipated and ground truth segmentations. The Dice Coefficient is calculated using the following formula:
\small{  
\begin{equation} \text{Dice}_{\text{Coeff}} = \frac{2 \times \text{Area of Intersection}}{\text{Area of Prediction} + \text{Area of Ground Truth}} 
\end{equation}
}
Here: 
\begin{itemize}
    \item \textbf{Area of Intersection} is the total area where the predicted segmentation and the ground truth segmentation overlap.
    \item \textbf{Area of Prediction} is the total area covered by the predicted segmentation.
    \item \textbf{Area of Ground Truth} is the total area covered by the ground truth segmentation. 
\end{itemize}
 \textbf{Average Precision (AP),}  \cite{AP}  for instance segmentation considers both the accuracy of bounding box predictions and the quality of pixel-level masks.  It is computed by comparing predicted masks to ground truth masks using IoU. The prediction is judged accurate if the bounding box and mask overlap sufficiently with the ground truth. The AP is calculated and averaged over a set of IoU criteria.

 \begin{itemize}
\item \textbf{\(AP_{IoU=0.5}\):} Average Precision calculated when considering predictions with an IoU of at least 0.5.
\item \textbf{\(AP_{IoU=0.75}\):} Average Precision calculated when considering predictions with an IoU of at least 0.75.
\end{itemize}

\subsubsection{Evaluation Metrics for Image Classification\newline } 
We evaluated our suggested system's classification capabilities using several matrices, including Accuracy, Precision, Recall, F1 Score, and Logarithmic Loss \cite{loss}. 

\textbf{Accuracy} represents the ratio of correctly categorized cases to the total number of cases. It assesses how well our model can classify potato diseases. The score can be computed as follows:
\newline

\begin{equation}
\text{Accuracy} =\displaystyle\frac{TP_{PD} + TN_{PD}}{TP_{PD} + TN_{PD} + FP_{PD} + FN_{PD}}
\end{equation}

Here:
\begin{itemize}
    \item $TP$ stands for true positive.
    \item $TN$ means true negative.
    \item $FP$ indicates false positive.
    \item $FN$ implies false negative.
    \item $PD$ denotes potato disease.
\end{itemize}

\textbf{Precision} is the proportion of true positives to the sum of true positives and false positives. In the context of our research, true positive signifies the right prediction of the existence of a specific potato disease, whereas false positive describes cases in which the model mistakenly predicts the presence of a disease when it does not exist. Precision is computed as follows: 
\newline

\begin{equation}
\text{Precision} = \displaystyle\frac{TP_{PD}}{TP_{PD} + FP_{PD}}
\end{equation}

\textbf{Recall } analyzes the model's ability to detect and categorize diseases into proper disease categories, ensuring that only a minimal amount of samples are ignored or misclassified. The formula for the recall is as follows:
\newline

\begin{equation}
\text{Recall} = \displaystyle\frac{TP_{PD}}{TP_{PD} + FN_{PD}}
\end{equation}

\textbf{F1 Score} is the harmonic mean of precision and recall. It is especially beneficial for balancing the trade-off between false positives and false negatives in disease detection. The score can be computed as follows:

\begin{equation}
\text{F1 Score} = \displaystyle\frac{2 \times \text{Precision}_{PD} \times \text{Recall}_{PD}}{\text{Precision}_{PD} + \text{Recall}_{PD}}
\end{equation}

\textbf{Logarithmic Loss}, commonly referred to as Log Loss, is a metric used to assess the performance of classification models in multiclass instances. It measures the accuracy of a model's anticipated class probabilities, rewarding correct and confident predictions while penalizing uncertain or incorrect ones. 

The logarithmic loss (Log Loss) for disease detection is calculated as follows:

\begin{equation}
\text{Log Loss} = -\frac{1}{N} \sum \left[ y \cdot \log(p) + (1 - y) \cdot \log(1 - p) \right]
\end{equation}

\text{Where:} \\
\begin{itemize}
    \item $N$: Total number of images.
    \item $y$: True label (1 if the image belongs to a specific potato disease, 0 otherwise).
    \item $p$: The predicted probability that the image belongs to the specific potato disease.
\end{itemize}

\section{Proposed Methodology} \label{Methodologylab}
\begin{figure*}
\centering
\includegraphics[width=470pt, height=230pt]{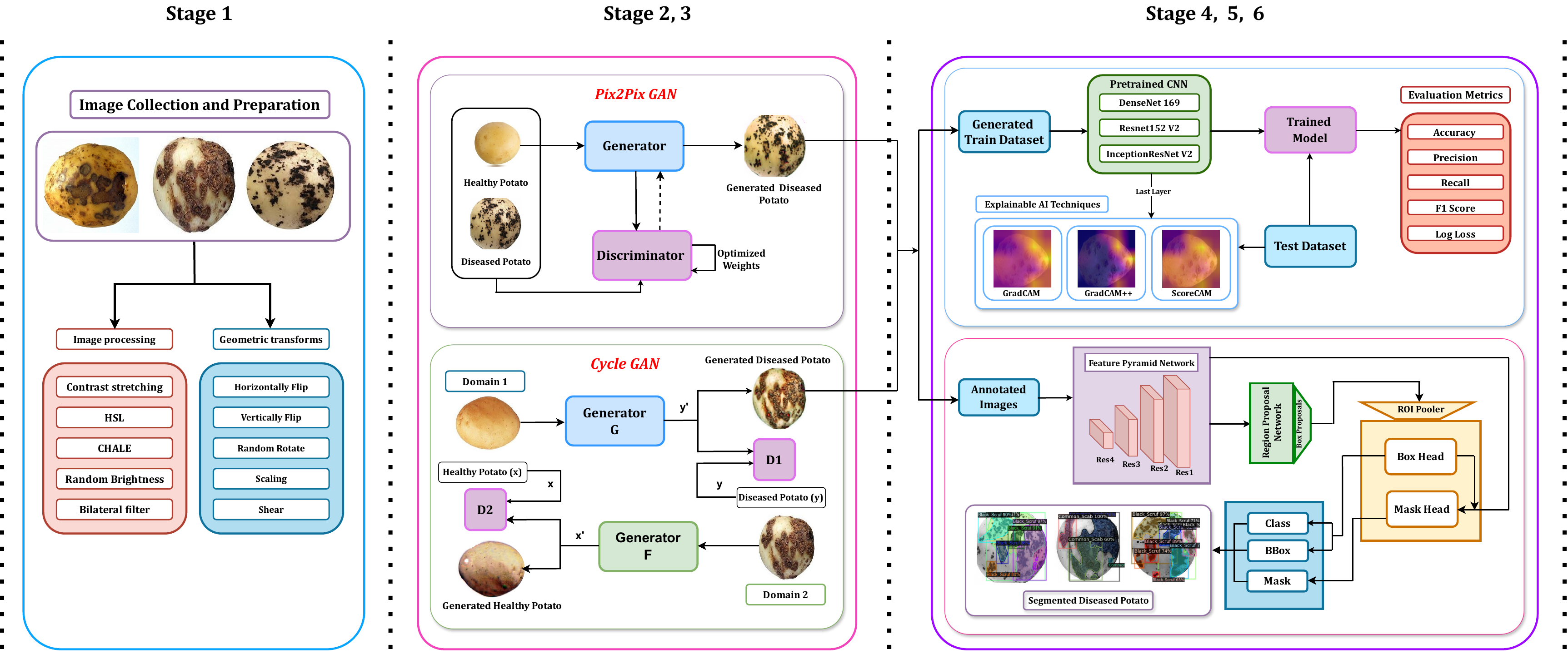}
\caption{Visual Diagram of our Proposed Potato Disease Image Generation and Disease Segmentation System, meticulously designed to capture and transform specific regions, providing a comprehensive grasp of the dynamic process from the creation of images to the accurate segmentation of diseases
}\label{fig:workflow}
\end{figure*}

This section describes the detailed process of our suggested technique, which guides the way you through the steps of Potato Disease Image Generation. Beginning with perfect health, several example images are presented as input, each of which goes through a changing process that has been meticulously recorded for demonstrative purposes. Random samples from the dataset are carefully chosen to demonstrate the richness and variety of our approach. Figure \ref{fig:workflow} is a visual narrative that depicts the methodological challenges of our research, from the creation of healthy images to the dynamic progression of disease image generation, culminating in the precision of disease Segmentation. The complexities of each stage are carefully explained, demonstrating the creative and thorough characteristics of our suggested technique.

\textbf{Stage 1) Input Image:}
In the initial stage, we standardize the images for analysis by resizing them to a consistent resolution of (224 $\times$ 224) pixels. All the steps for preparing the input image for training are outlined in the section \ref{Datasetlab}. We rigorously follow these steps before feeding the images into GAN models.

\textbf{Stage 2) Image Translation with CycleGAN and Pix2Pix:}
At this stage, we begin implementing the image translation process using both the CycleGAN and Pix2Pix GAN models. These models are first developed with the specific goal of translating healthy potato images into real disease-like counterparts. We proceed to train both models using a training dataset that contains labeled healthy and diseased potato images. During the training phase, hyperparameters and model architectures are adjusted as needed to maximize overall performance. Following training, the models are evaluated using a separate testing dataset, allowing for an in-depth assessment of their effectiveness and generalization capabilities.

\textbf{Stage 3) Assessment of Realistic Disease Image Generation:}
During this stage, we apply the Inception Score and Fréchet Inception Distance (FID) to analyze the realism of the generated disease images. We calculate the Inception Score to assess the diversity and quality of the generated images. Furthermore, we compute the FID to analyze the similarity of the distributions of actual and created images.

\textbf{Stage 4) Explainable AI for Interpretability in Potato Disease Classification:}
In the fourth stage of our research, we want to enhance the interpretability of the potato disease classification. This involves the easy incorporation of three cutting-edge gradient-based XAI techniques: GradCAM, GradCAM++, and ScoreCAM. Meanwhile, we work with three other CNN architectures: DenseNet 169, ResNet152 V2, and InceptionResNet V2. Our novel method aims to provide deep insights into the decision-making processes of neural networks by providing informative and visually stimulating representations. Our major objective is to uncover the underlying mechanisms that influence the classification of potato illnesses using a complex combination of advanced visualization techniques and a broad variety of convolutional neural network models.

\textbf{Stage 5) Detectron2 Configuration:}
We configure Detectron2 at this stage to achieve specific potato disease segmentation, utilizing the capabilities of Mask R-CNN and experimenting with ResNet50, ResNet101, and ResNeXt-101 backbones. The system we have developed has been thoroughly built to interface with our dataset, in which we specify distinct categories for healthy and diseased potatoes. We then fine-tune the pre-trained models using our custom-built dataset to achieve precise and effective segmentation of potato diseases.

\textbf{Stage 6) Performance Assessment of Potato Disease Classification and Segmentation:}
We assessed our proposed system's classification performance using a variety of metrics, including accuracy, recall, precision, F1 score, and Log Loss. To evaluate potato disease segmentation at this stage, we rely on Detectron2 predictions on the testing set. We measure the overlap between predicted and ground truth segmentation masks using Average Precision (AP) over different Intersection Over Union (IoU) and Dice Score metrics. The IOU measures the intersection area to the union area ratio. In contrast, the Dice Score computes twice the intersection area divided by the sum of areas, offering quantitative insights on segmentation accuracy and the model's capacity to detect damaged potato areas.

\section{Result Analysis}\label{resultlab}
\subsection{Experimental Setup}
The research was conducted on Jupyter Notebook and Kaggle. The PyTorch version used was V2.1.0. The GPUs used were NVIDIA GeForce RTX 3050 with 8.6 Compute Capability and Tesla T4 GPU with 7.5 Compute Capability. The CPUs used were an Intel Core i5 9400f with a maximum clock speed of 4.25 GHz and an Intel(R) Xeon(R) CPU @ 2.20GHz. The RAM used was 16 GB and 30 GB.

\subsection{Hyper-parameter settings}

\begin{table*}
\caption{Optimizing Hyperparameters for Realistic Disease Image Generation}\label{table3}
\centering
  \begin{tabular}{lllllll} 
    \toprule
 \textbf{Class} & \textbf{Models} & \textbf{Learning Rate} & \textbf{Batch Size} & \textbf{Number of Epochs} & \textbf{Optimizer}  \\
    \midrule
    {Black Scurf} & Cycle GAN & \(1e\)\(^{-5}\) & 8 & 70 & Adam   \\
       & Pix2Pix GAN  & \(2e\)\(^{-4}\) & 8 & 130 & Adam   \\
     \\
   {Common Scab} & Cycle GAN  & \(1e\)\(^{-5}\) & 8 & 70 & Adam  \\
        & Pix2Pix GAN & \(2e\)\(^{-4}\) & 8 & 130 & Adam \\
    \bottomrule 
  \end{tabular}
\end{table*}

\begin{table}
\caption{Optimizing Hyperparameters for Potato Disease Instance Segmentation using the Detectron2 Framework Baselines with Mask R-CNN}\label{table4}
\centering
  \begin{tabular}{p{2.8cm} p{2.1cm} p{1.5cm} p{2.2cm} p{2.4cm}}
    \toprule
 \textbf{Backbone}  & \textbf{Learning Rate} & \textbf{Batch \newline Size} & \textbf{Number of Epochs} & \textbf{Optimizer}  \\
    \midrule
    {ResNet-50} & \(1e\)\(^{-3}\) &  8 & 25 & SGD    \\
     
   {ResNet-101}   & \(1e\)\(^{-3}\) & 8 & 25 & SGD   \\ 
     
      {ResNeXt-101}  & \(1e\)\(^{-3}\) & 8 & 25 & SGD   \\  
    
    \bottomrule 
  \end{tabular}
\end{table}

\begin{table}
\caption{Hyperparameter Tuning for Potato Disease Classification Using Pretrained CNNs}\label{table5}
\centering
  \begin{tabular}{lllll}
    \toprule
 \textbf{Models}  & \textbf{Learning Rate} & \textbf{Batch Size} & \textbf{Number of Epochs} & \textbf{Optimizer}  \\
    \midrule
    {DenseNet169} & \(1e\)\(^{-2}\) & 10 & 30 & Adam    \\
     
   {Resnet152V2}   & \(1e\)\(^{-2}\) & 10 & 30 & Adam   \\ 
     
      {InceptionResNetV2}  & \(1e\)\(^{-2}\) & 10 & 30 & Adam   \\  
    
    \bottomrule 
  \end{tabular}
\end{table} 

Table \ref{table3} presents the optimal hyperparameter settings for two GAN models CycleGAN and Pix2Pix GAN used to generate realistic images of two potato diseases: Black Scurf and Common Scab. CycleGAN performs well with a learning rate of \(1e\)\(^{-5}\), batch size of 8, and 70 epochs, while Pix2Pix GAN requires a higher learning rate of \(2e\)\(^{-4}\) and longer training 130 epochs.
 
Table \ref{table4} shows the hyperparameters for training Mask R-CNN models on potato disease segmentation using Detectron2 with different backbones, such as ResNet-50, 101, and ResNetXt-101. The same batch size of 8, epochs of 25, learning rate of 1e-3, and SGD optimizer were used in all models.

Table \ref{table5} details the hyperparameters used to train various pre-trained CNN models for potato disease classification, such as DenseNet169, Resnet152V2, and InceptionResNetV2. All models were trained with identical settings: a batch size of 10, 30 epochs, a learning rate of 1e-2, and the Adam optimizer.

\begin{figure}
    \centering
    \includegraphics[width=95mm,scale=2]{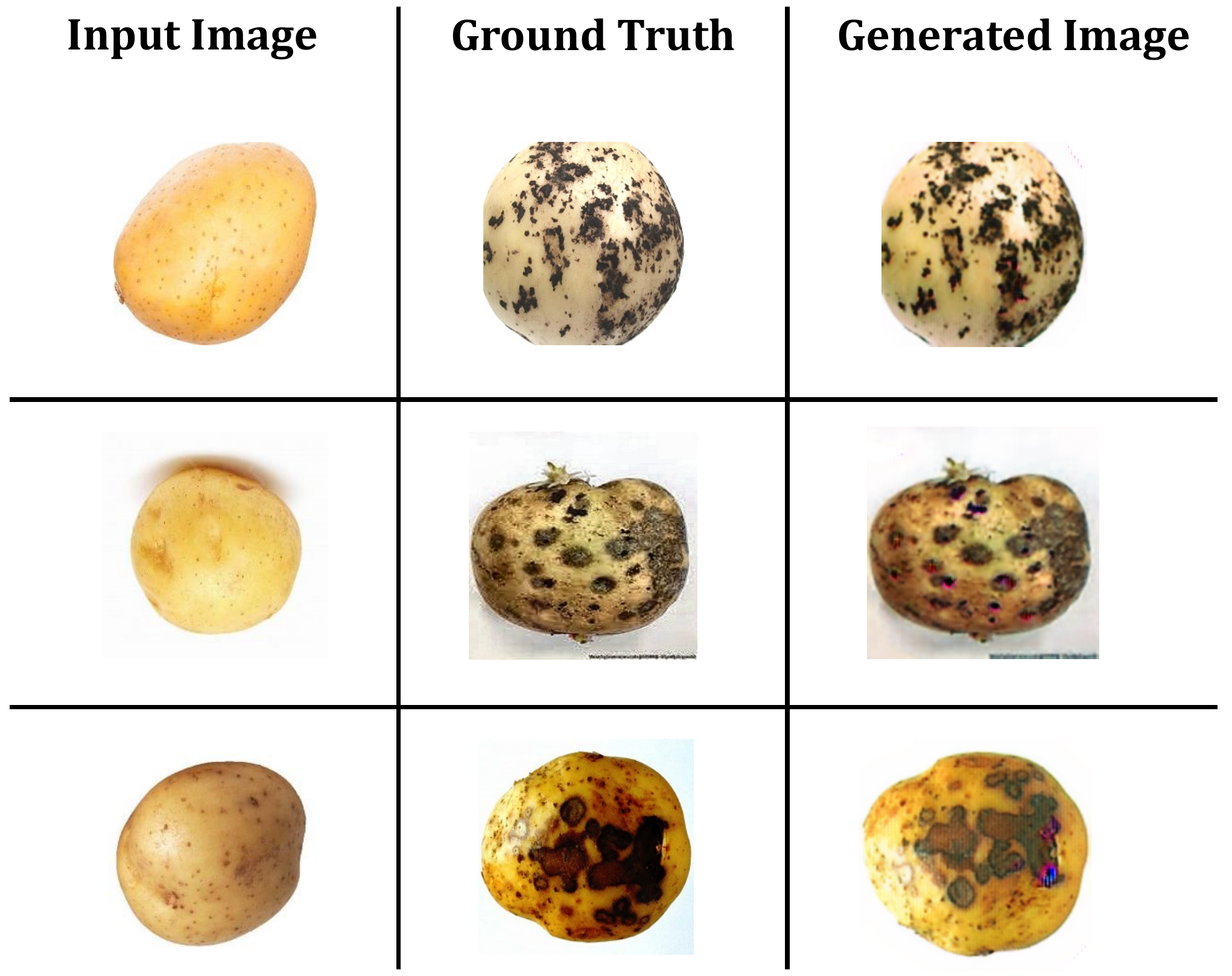}
    \caption{Visual Representation of Generated Realistic Potato Disease Images}
    \label{fig:Comparison}
\end{figure}

\begin{figure*}
    \centering
    \begin{subfigure}{0.32\textwidth}
        \includegraphics[width=\linewidth]{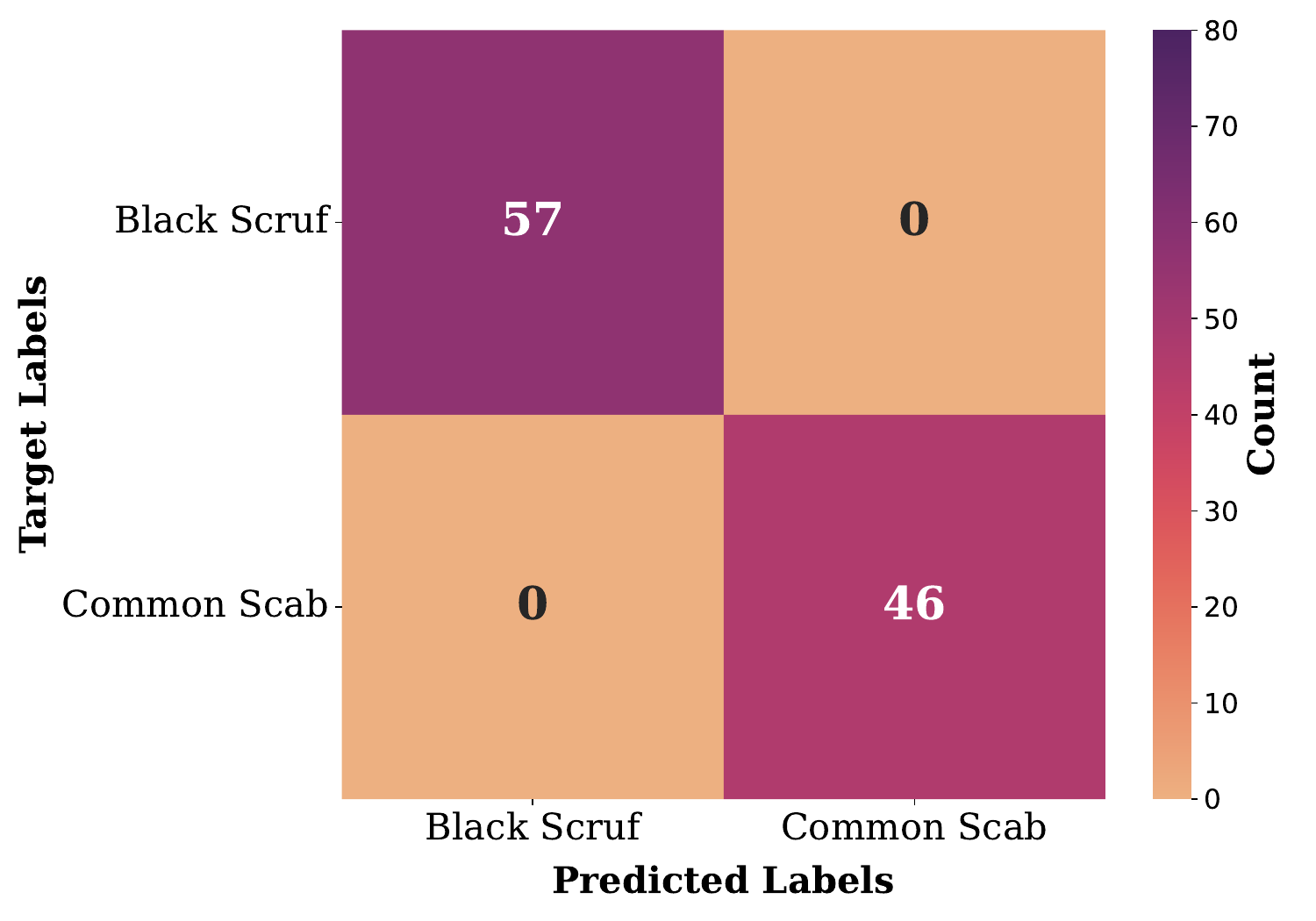}
        \caption{DenseNet169}
        \label{fig:subC1}
    \end{subfigure}
    \hfill
    \begin{subfigure}{0.32\textwidth}
        \includegraphics[width=\linewidth]{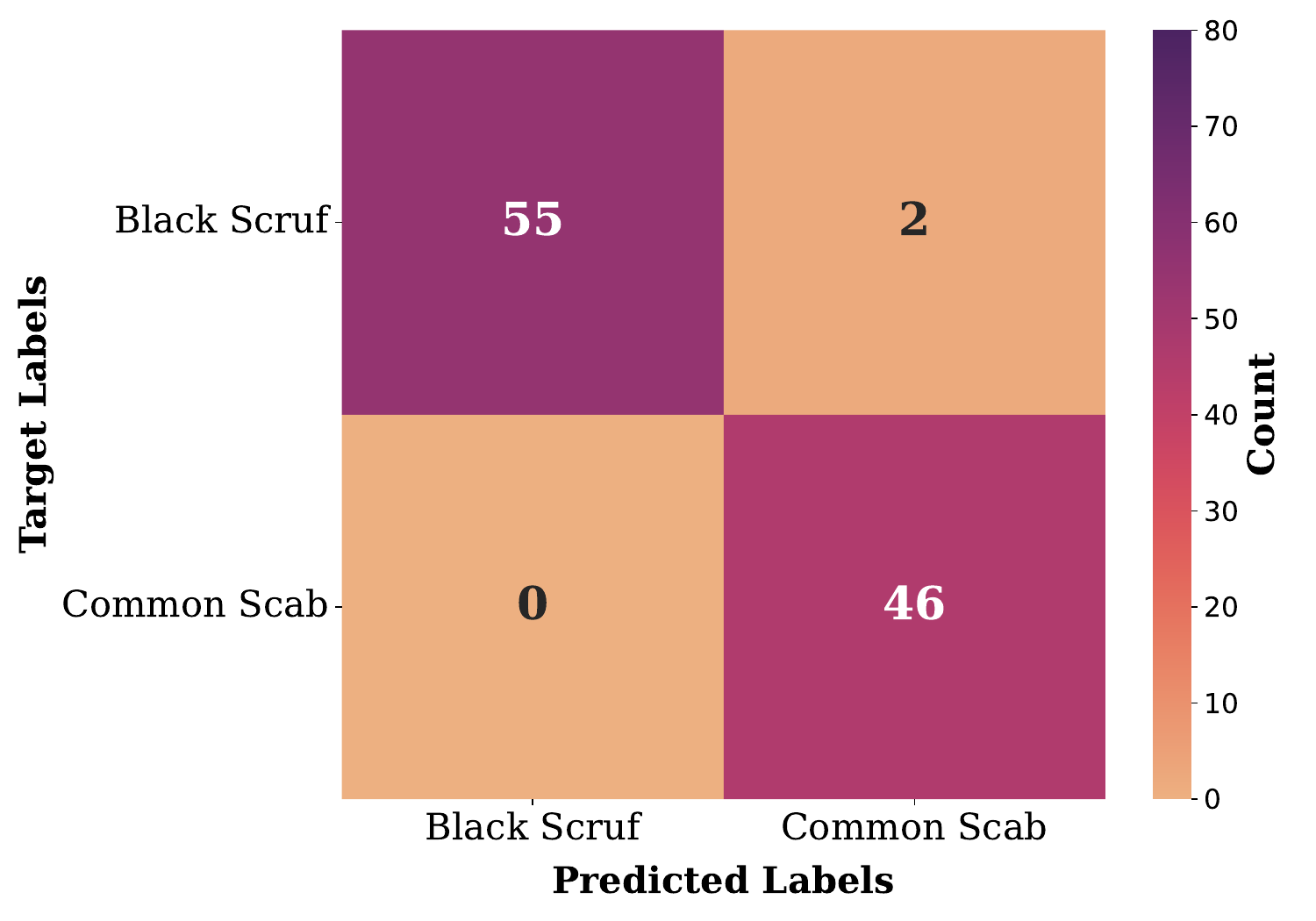}
        \caption{Resnet152V2} 
        \label{fig:subC2}
    \end{subfigure}
    \hfill
    \begin{subfigure}{0.32\textwidth}
        \includegraphics[width=\linewidth]{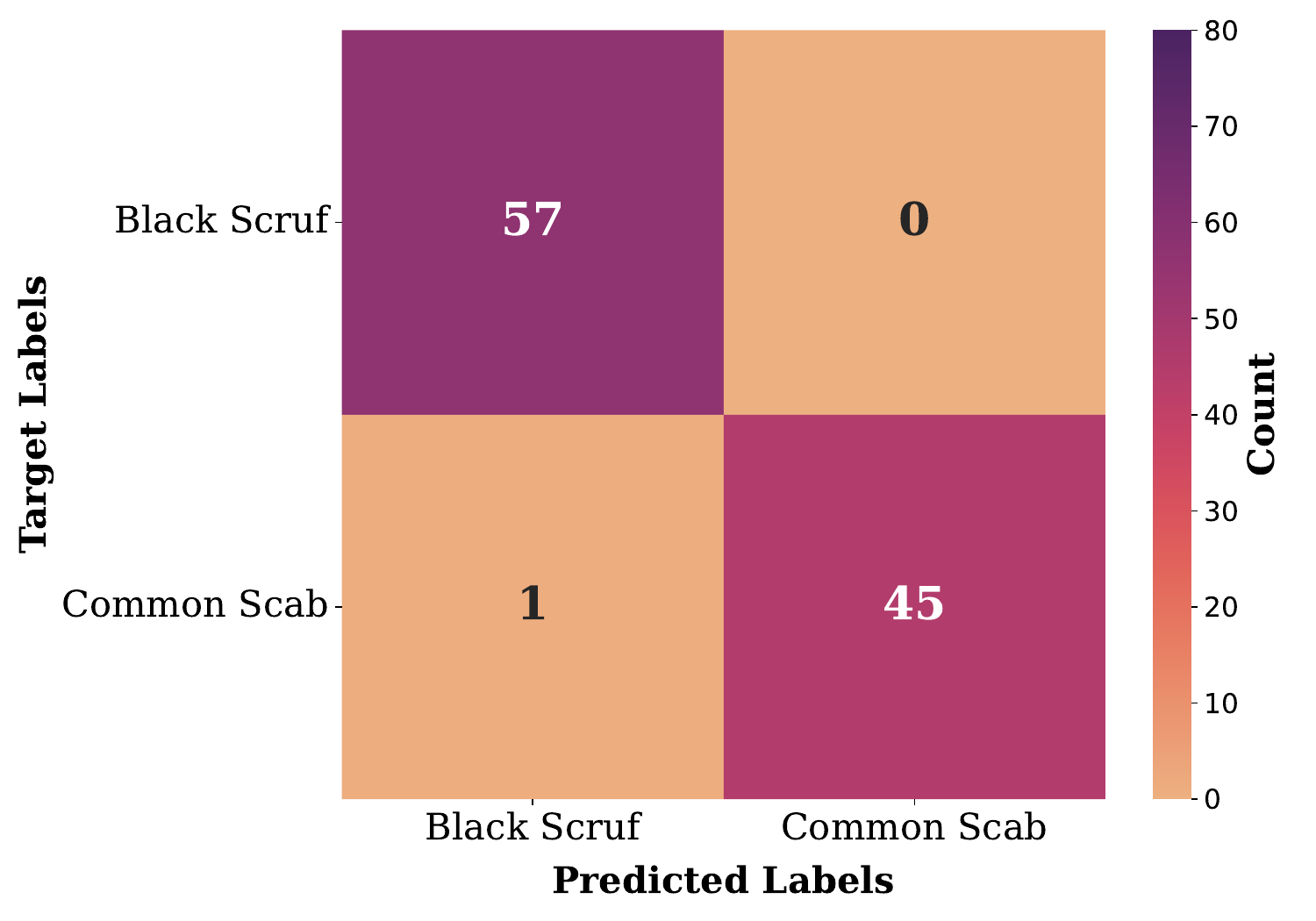}
        \caption{InceptionResNetV2}
        \label{fig:subC3}
    \end{subfigure}

    \caption{Comparison of Confusion Matrices from Three Pretrained CNNs for Potato Disease Classification}\label{fig:CM}

\end{figure*}

\subsection{Results}

\begin{figure}
    \centering
    \begin{subfigure}{0.32\textwidth}
        \includegraphics[width=\linewidth]{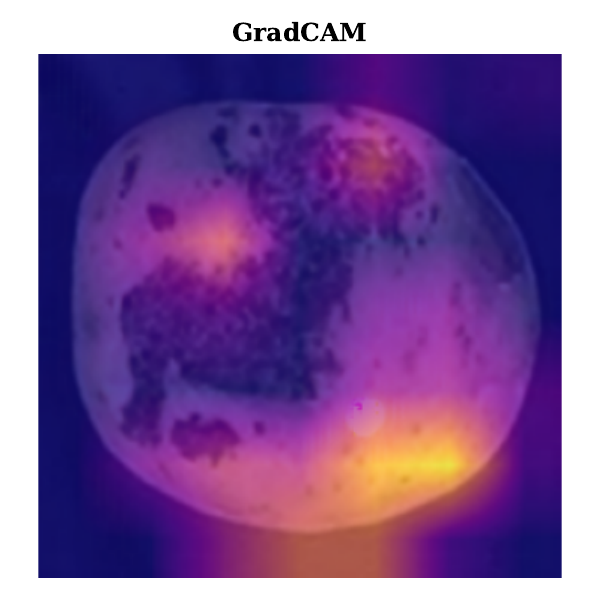}
        \caption{DenseNet169: GradCAM}
        \label{fig:sub1}
    \end{subfigure}
    \hfill
    \begin{subfigure}{0.32\textwidth}
        \includegraphics[width=\linewidth]{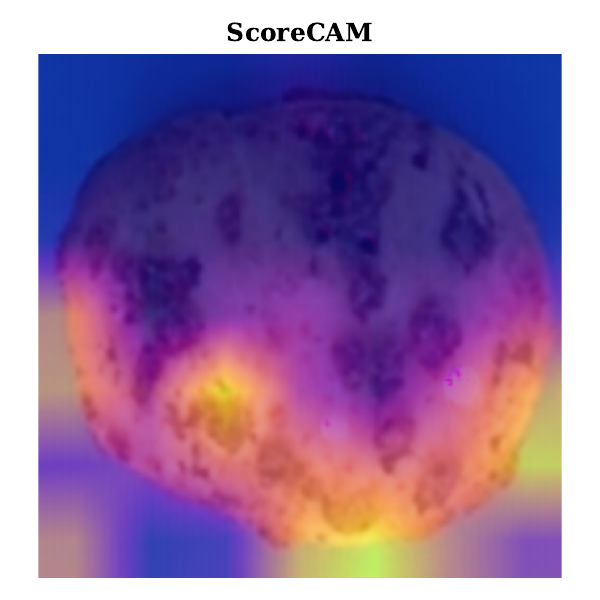}
        \caption{DenseNet169: ScoreCAM} 
        \label{fig:sub2}
    \end{subfigure}
    \hfill
    \begin{subfigure}{0.32\textwidth}
        \includegraphics[width=\linewidth]{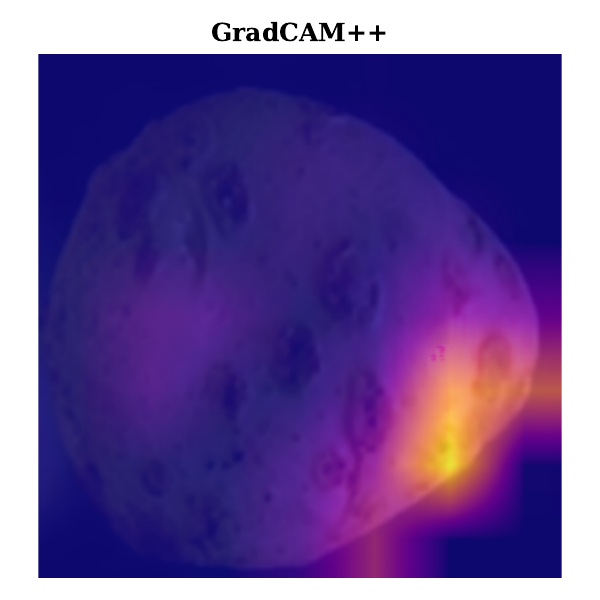}
        \caption{DenseNet169: GradCAM++}
        \label{fig:sub3}
    \end{subfigure}

    \medskip
    
    \begin{subfigure}{0.32\textwidth}
        \includegraphics[width=\linewidth]{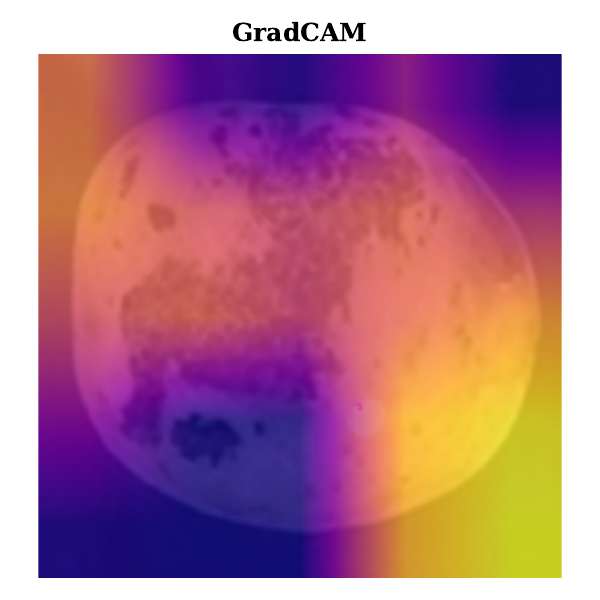}
        \caption{InceptionResNetV2: GradCAM}
        \label{fig:sub4}
    \end{subfigure}
    \hfill
    \begin{subfigure}{0.32\textwidth}
        \includegraphics[width=\linewidth]{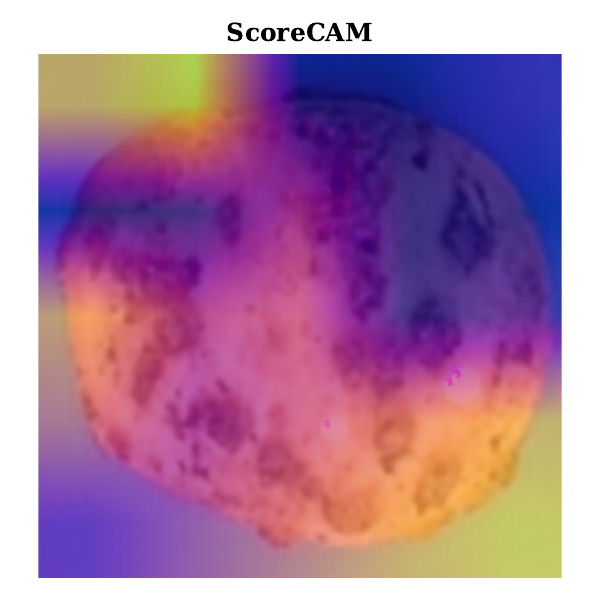}
        \caption{InceptionResNetV2: ScoreCAM} 
        \label{fig:sub5}
    \end{subfigure}
    \hfill
    \begin{subfigure}{0.32\textwidth}
        \includegraphics[width=\linewidth]{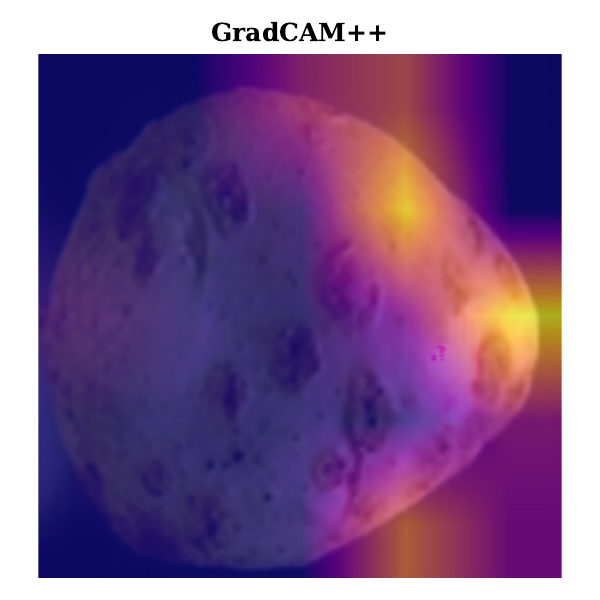}
        \caption{InceptionResNetV2: GradCAM++}
        \label{fig:sub6}
    \end{subfigure}

    \medskip
    
    \begin{subfigure}{0.32\textwidth}
        \includegraphics[width=\linewidth]{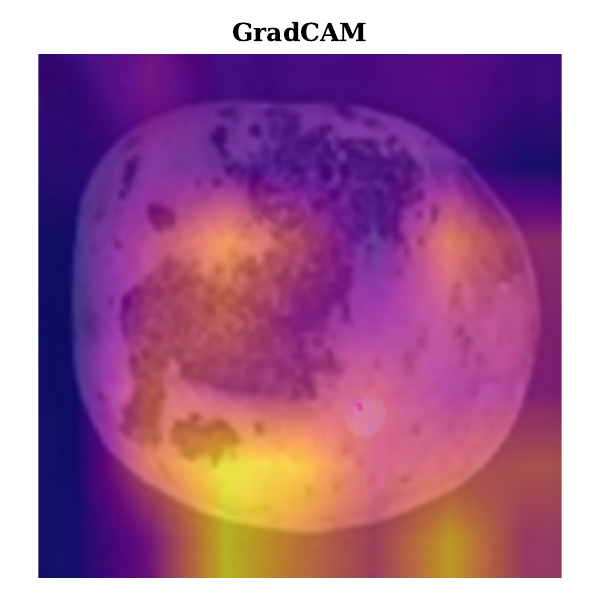}
        \caption{ResNet152V2: GradCAM}
        \label{fig:sub7}
    \end{subfigure}
    \hfill
    \begin{subfigure}{0.32\textwidth}
        \includegraphics[width=\linewidth]{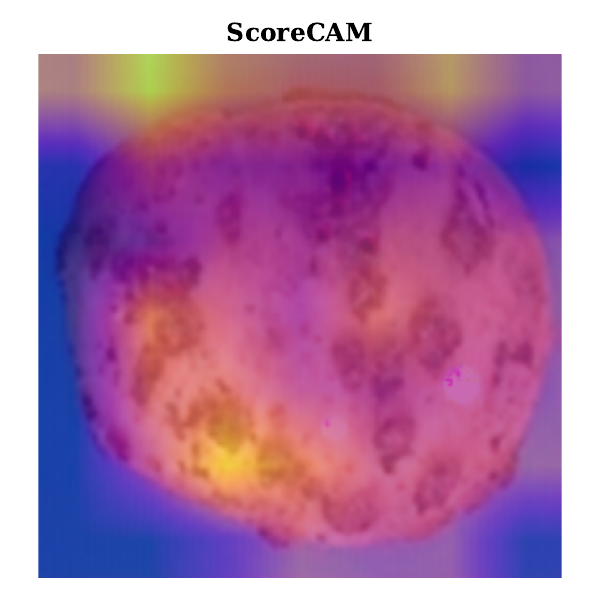}
        \caption{ResNet152V2: ScoreCAM} 
        \label{fig:sub8}
    \end{subfigure}
    \hfill
    \begin{subfigure}{0.32\textwidth}
        \includegraphics[width=\linewidth]{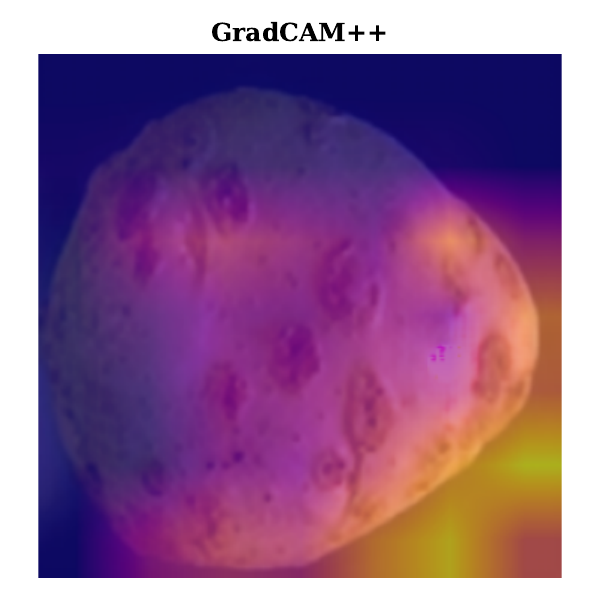}
        \caption{ResNet152V2: GradCAM++}
        \label{fig:sub9}
    \end{subfigure}
    
    \caption{Insights into Potato Disease Classification from an Explainable AI Perspective} 
    \label{fig:nineimages} 

\end{figure}

\begin{figure}
    \centering
    \includegraphics[width=95mm,scale=2]{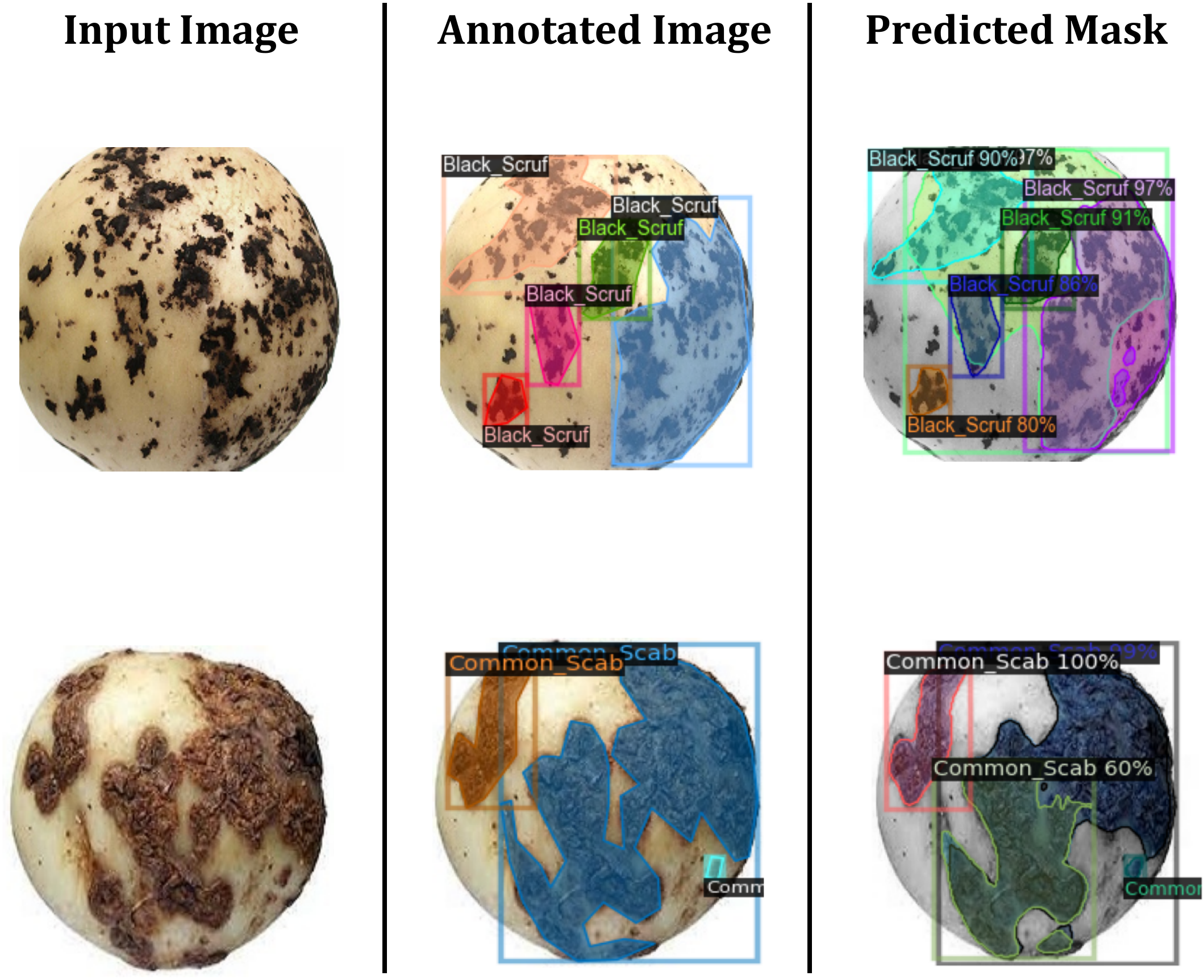}
    \caption{Visual Representation of Segmented Potato Disease Images} \label{fig:seg}
\end{figure}

The performance of two Generative Adversarial Network (GAN) models, CycleGAN and Pix2Pix GAN, in the generation of realistic images representing potato diseases, is compared in Table \ref{GANtable}. The evaluation makes use of two metrics: Frechet Inception Distance (FID) and Inception Score (IS). CycleGAN outperforms Pix2Pix GAN in both metrics across both potato disease categories, according to the results. In the case of black scurf, for example, CycleGAN has a lower FID score of 0.4028 compared to Pix2Pix GAN's 0.5743, but a higher IS score of 1.2001 compared to Pix2Pix GAN's 0.9899. This suggests that CycleGAN is better than Pix2Pix GAN at generating realistic and varied images of black scurf. Similarly, for Common Scab, CycleGAN has a lower FID score of 0.4882 than Pix2Pix GAN, but a higher IS score of 1.0900 than Pix2Pix GAN. This suggests that CycleGAN outperforms Pix2Pix GAN in producing more realistic and diverse images of Common Scab. Visualization of Generated Realistic Potato Disease is shown in Figure \ref{fig:Comparison}.

Table \ref{tableClass} presents the performance evaluation of pretrained convolutional neural network (CNN) models for potato disease classification. Three models DenseNet169, ResNet152V2, and InceptionResNetV2 are compared based on five metrics: Accuracy, Precision, Recall, F1 Score, and Log Loss. DenseNet169 Achieved perfect scores (1.0) in Accuracy, Precision, Recall, and F1 Score, indicating excellent performance in correctly classifying instances of potato diseases. It also showed the lowest Log Loss 0.0024   among the models. ResNet152V2 and InceptionResNetV2 Both models performed well but with slightly lower scores compared to DenseNet169 across all metrics. ResNet152V2 achieved an accuracy of 0.9804, with precision and recall scores of 0.9792 and 0.9821, respectively. The F1 score, which balances precision and recall, reached 0.9803. However, its log loss was relatively higher at 0.7067. On the other hand, InceptionResNetV2 exhibited slightly superior performance, achieving an accuracy of 0.9902 and precision and recall scores of 0.9912 and 0.9891, respectively. The F1 score for InceptionResNetV2 was 0.9901, indicating a robust balance between precision and recall. Additionally, its log loss was notably lower at 0.3533. The Confusion matrices of Pretrained CNNs  are shown in Figure \ref{fig:CM}

Figure \ref{fig:nineimages} presents a comparative analysis of Three explainable AI techniques— GradCAM, GradCAM++, and ScoreCAM. These techniques are applied to three diverse convolutional neural networks (CNNs) DenseNet169, InceptionResNetV2, and ResNet152. These techniques highlight which parts of an image influence CNN's decision-making process for a specific class. By visually examining the output visualizations produced by each technique across different CNN architectures. Based on these figures, GradCAM performs mediocrely overall. we can see that GradCAM++ appears to highlight the most relevant image regions, particularly when used with DenseNet169 or InceptionResNetV2. Conversely, ResNet152 with all three explainable AI techniques seems to focus on a lot of irrelevant areas in its decision-making process. 

        
\begin{table}
\caption{Evaluation of Realistic Potato Disease Images Generated via Frechet Inception Distance and Inception Score}\label{GANtable}
\centering
  \begin{tabular}{lllll} 
    \toprule
 \textbf{Class} & \textbf{GANs}&
       \textbf{\textbf{Frechet Inception Distance}}&
       \textbf{\textbf{Inception Score}}\\
    \midrule
   \textbf{Black Scurf} &  Cycle GAN  & \hspace{38pt} 0.4028 & \hspace{18pt} 1.2001  \\
     &  Pix2Pix GAN  & \hspace{38pt} 0.5743 & \hspace{18pt} 0.9899 \\
     \midrule
      \textbf{Common Scab} &  Cycle GAN & \hspace{38pt} 0.4882 & \hspace{18pt}  1.0900      \\
      &  Pix2Pix GAN & \hspace{38pt} 0.6240 & \hspace{18pt} 0.9643\\ 
    
    \bottomrule
  \end{tabular}
\end{table}
\begin{table}
\caption{Assessing the Performance of Pretrained CNNs in Potato Disease Classification}\label{tableClass}
\centering
  \begin{tabular}{p{3.0cm} p{1.5cm} p{1.5cm} p{1.5cm} p{1.5cm}p{1.5cm}}
    \toprule
 \textbf{Model}  & \textbf{Accuracy} & \textbf{Precision} & \textbf{Recall} & \textbf{F1 Score} & \textbf{Log Loss}  \\
    \midrule
    {DenseNet169} & 1.0000 & 1.0000 & 1.0000 & 1.0000 &  0.0024   \\
     
   {Resnet152V2}   & 0.9804 & 0.9792 & 0.9821 & 0.9803 & 0.7067  \\ 
     
      {InceptionResNetV2}  & 0.9902 & 0.9912 & 0.9891 & 0.9901 &  0.3533\\  
    
    \bottomrule 
  \end{tabular}
\end{table}

\begin{table}
\caption{Assessment of Potato Disease Instance Segmentation Performance Utilizing Detectron2 COCO Instance Segmentation Baselines\\}\label{DetectronTab}
\centering
  \begin{tabular}{lllllll} 
    \toprule
 \textbf{Backbone} & \textbf{Task Type}&
       \textbf{\textbf{\hspace{10pt} AP}}&
       \textbf{\textbf{\(AP_{IoU= 0.5}\)}}&
       \textbf{\textbf{\(AP_{IoU= 0.75}\)}}&
       \textbf{\textbf{Dice Score}} \\
    \midrule
   ResNet-50 & Segmentation  & \hspace{5pt} 73.204 & \hspace{5pt} 89.733 & \hspace{5pt} 86.126 & \hspace{5pt} 0.6014 \\
     & Bounding Box  & \hspace{5pt} 83.824 & \hspace{5pt} 90.526 & \hspace{5pt} 86.353 & \\
    \midrule
      ResNet-101 & Segmentation  & \hspace{5pt} 78.681 & \hspace{5pt} 92.905 & \hspace{5pt} 74.851 & \hspace{5pt} 0.6728 \\
     & Bounding Box  & \hspace{5pt} 87.886 & \hspace{5pt} 96.409 & \hspace{5pt} 90.943 & \\
    \midrule
    ResNeXt-101 & Segmentation  & \hspace{5pt} 86.039 & \hspace{5pt} 97.030 & \hspace{5pt} 96.040 & \hspace{5pt} 0.8112 \\
     & Bounding Box  & \hspace{5pt} 97.030 & \hspace{5pt} 97.030 & \hspace{5pt} 97.030 & \\
    \bottomrule
  \end{tabular}
\end{table}

Table \ref{DetectronTab} gives a complete evaluation of Potato Disease Instance Segmentation using several Mask R-CNN backbones. The metrics that have been evaluated include Average Precision (AP), \(AP_{IoU= 0.5}\), \(AP_{IoU= 0.75}\), and the Dice Score.
The ResNet-50 backbone performs well in the Segmentation task, with an AP of 73.204, \(AP_{IoU= 0.5}\) of 89.733, \(AP_{IoU= 0.75}\) of 86.126, and a Dice Score of 0.6014. In the Bounding Box task, ResNet-50 has an AP of 83.824, \(AP_{IoU= 0.5}\) of 90.526, and \(AP_{IoU= 0.75}\) of 86.353. Moving on to ResNet-101 as the backbone, in the Segmentation task, it records an AP of 78.681, \(AP_{IoU= 0.5}\) of 92.905, \(AP_{IoU= 0.75}\) of 74.851, and a Dice Score of 0.6728. ResNet-101 obtains an AP of 87.886, \(AP_{IoU= 0.5}\) of 96.409, and \(AP_{IoU= 0.75}\) of 90.943 on the Bounding Box task. The third backbone, ResNeXt-101, outperforms in both Segmentation and Bounding Box challenges. The Segmentation task results in an AP of 86.039, \(AP_{IoU= 0.5}\) of 97.030, \(AP_{IoU= 0.75}\) of 96.040, and a high Dice Score of 0.8112. ResNeXt-101 scores the highest on the Bounding Box task, with an AP of 97.030, \(AP_{IoU= 0.5}\) of 97.030, and \(AP_{IoU= 0.75}\) of 97.030. 
In summary, the results indicate that ResNeXt-101 beats ResNet-50 and ResNet-101 on all parameters, demonstrating its usefulness in Potato Disease Instance Segmentation tasks. Furthermore, the Dice Score gives useful information on segmentation performance, with ResNeXt-101 showing greater overlap between predicted and ground truth masks. Visual depictions of segmented potato disease regions are shown in Figure \ref{fig:seg}.

\section{Future Directions in Precision Agriculture}\label{futurelab}
In the realm of agricultural research, we envision a future where our endeavors extend beyond the familiar confines of potato cultivation, embracing the wide range of crops that sustain our global food supply. Our next research projects will focus on a comprehensive investigation of diseases that are particular to crops, to create customized strategies that effectively and precisely tackle these problems. The investigation of innovative approaches for precise volume estimates is a key component of our future research agenda. The combination of computer vision and machine learning technologies makes this task feasible. Although we have initially concentrated on potatoes, our diversification into other crops demonstrates our dedication to improving agricultural methods in their entirety. Our goal is to accurately estimate crop volumes in the field by utilizing cutting-edge algorithms. This will provide invaluable insights for yield estimations, resource optimization, and strategic harvest planning of different crops. Furthermore, we intend to further explore the field of eXplainable Artificial Intelligence (XAI) to improve and enhance the interpretability of our models. Our dedication to openness and comprehension will serve as the foundation for our attempts to combat agricultural diseases more precisely and effectively.

\section{Conclusion}\label{conclusionlab}
Potato crops play a significant role in securing worldwide food security. Recognizing the need for effective disease prevention and management in potato agriculture, our study offers a novel technique called PotatoGANs. In this innovative technique, we use two types of Generative Adversarial Networks (GANs) to create synthetic images of potato diseases, with a focus on Common Scab and Black Scruf. This method differs from standard data augmentation strategies in that it allows us to generate a unique dataset. In our research, we found that integrating Generative Adversarial Networks (GANs), CycleGAN outperforms Pix2Pix GAN in image quality for black scurf and common scab, with lower FID scores of 0.4028 and 0.4882 compared to Pix2Pix GAN 0.5743 and 0.6240. CycleGAN also produces more diverse images, evident in higher IS scores of 1.2001 and 1.0900 compared to Pix2Pix GAN 0.9899 and 0.9643. 
Our model performs exceptionally well in the field of classifying potato diseases using a variety of CNN architectures. With an accuracy of 1.0000, DenseNet169 reaches perfection, ResNet152 V2 performs robustly with an accuracy of 0.9803, and InceptionResNet V2 achieves remarkable distinction with an accuracy of 0.9901. Furthermore, we used Detectron 2 for Potato Disease Segmentation, which produced excellent results in localizing damaged areas. With maximum scores of 86.039 for Average Precision and 0.8112 for the Dice Coefficient in the ResNeXt-101 backbone, we demonstrate the precision and efficacy of our disease segmentation approach. These findings highlight modern technology' breakthrough potential in improving potato detection and control, hence contributing to the development of sustainable agriculture.

\section*{Declarations}
\subsection*{Ethical Approval and Consent to participate}
Not Applicable.

\subsection*{Human and Animal Ethics}
Not Applicable.

\subsection*{Conﬂicts of Interest}
The authors declare that they have no conflicts of interest.

\subsection*{Funding}
This research was carried out with no external funding.
\subsection*{Availability of Data and Material}
The dataset used in this research is publicly accessible through \href{https://github.com/Wasi34/Comprehensive-Potato-Disease-Dataset}{GitHub}\footnote{\href{https://github.com/Wasi34/Comprehensive-Potato-Disease-Dataset}{https://github.com/Wasi34/Comprehensive-Potato-Disease-Dataset}}
\subsection*{Code Availability}
This research's codes and associated coding files are publicly available online on the \href{https://github.com/Mukaffi28/ExplainableAI-PotatoGAN-Cutting-Edge-Disease-Identification-for-Potatoes}{GitHub}\footnote{\href{https://github.com/Mukaffi28/ExplainableAI-PotatoGAN-Cutting-Edge-Disease-Identification-for-Potatoes}{https://github.com/Mukaffi28/ExplainableAI-PotatoGAN-Cutting-Edge-Disease-Identification-for-Potatoes}} platform. The codes are accessible to the public.

\subsection{Authors’ contributions}
Faria and Mukaffi outlined the research scope, conducted the study, gathered data, performed coding, led the majority of experiments, and drafted the manuscript. Wase aided in data collection, conducted multiple experiments, assessed writing quality, and rectified grammatical errors. Rabius contributed to both data collection and analysis. Alam provided essential supervision and guidance throughout the research, shaping its direction. Hasib provided essential supervision and critical editing for the manuscript.

\bibliographystyle{unsrt}
\bibliography{Mukaffi}

\end{document}